\documentclass[journal]{IEEEtran}

\IEEEoverridecommandlockouts
\usepackage{cite}
\usepackage{amsmath,amssymb,amsfonts}
\usepackage{algorithmic}
\usepackage{graphicx}
\usepackage{fixltx2e}
\usepackage{textcomp}
\usepackage{xcolor}
\usepackage{graphics}
\usepackage{subcaption}
\usepackage{caption}
\usepackage{float}
\usepackage{booktabs}
\usepackage{authblk}
\usepackage{array}
\usepackage{multirow}
\usepackage{mathtools}
\def\BibTeX{{\rm B\kern-.05em{\sc i\kern-.025em b}\kern-.08em
    T\kern-.1667em\lower.7ex\hbox{E}\kern-.125emX}}

\begin{document}

\title{Offloading Deep Learning Powered Vision Tasks from UAV to 5G Edge Server with Denoising}

\author[a]{Sedat Ozer 
\thanks{Corresponding Author: Sedat Ozer, e-mail: sedat.ozer@ozyegin.edu.tr \\ This is the preprint version. A version of this manuscript is accepted for publication at IEEE Transactions on Vehicular Technology.}}
\author[b]{Enes Ilhan}
\author[a,c]{Mehmet Akif Ozkanoglu}
\author[b]{Hakan Ali Cirpan}

\affil[a]{Ozer Lab, Department of Computer Science, Ozyegin University, Istanbul, Turkiye}
\affil[b]{Department of Electronic and Telecommunication Engineering, Istanbul Technical University, Istanbul, Turkiye}
\affil[c]{Ozer Lab, Department of Computer Engineering, Bilkent University, Ankara, Turkiye}

\maketitle

\begin{abstract}

Offloading computationally heavy tasks from an unmanned aerial vehicle (UAV) to a remote server helps improve the battery life and can help reduce resource requirements. Deep learning based state-of-the-art computer vision tasks, such as object segmentation and object detection, are computationally heavy algorithms, requiring large memory and computing power. Many UAVs are using (pretrained) off-the-shelf versions of such algorithms. Offloading such power-hungry algorithms to a remote server could help UAVs save power significantly. However, deep learning based algorithms are susceptible to noise, and a wireless communication system, by its nature, introduces noise to the original signal. When the signal represents an image, noise affects the image. There has not been much work studying the effect of the noise introduced by the communication system on pretrained deep networks. In this work, we first analyze how reliable it is to offload deep learning based computer vision tasks (including both object segmentation and detection) by focusing on the effect of various parameters of a 5G wireless communication system on the transmitted image and demonstrate how the introduced noise of the used 5G wireless communication system reduces the performance of the offloaded deep learning task. Then solutions are introduced to eliminate (or reduce) the negative effect of the noise. The proposed framework starts with introducing many classical techniques as alternative solutions first, and then introduces a novel deep learning based solution to denoise the given noisy input image. The performance of various denoising algorithms on offloading both object segmentation and object detection tasks are compared. Our proposed deep transformer-based denoiser algorithm (NR-Net) yields the state-of-the-art results on reducing the negative effect of the noise in our experiments.

\end{abstract}


\section{Introduction}

\begin{figure}[t]
    \centerline{\includegraphics[width=0.99\linewidth]{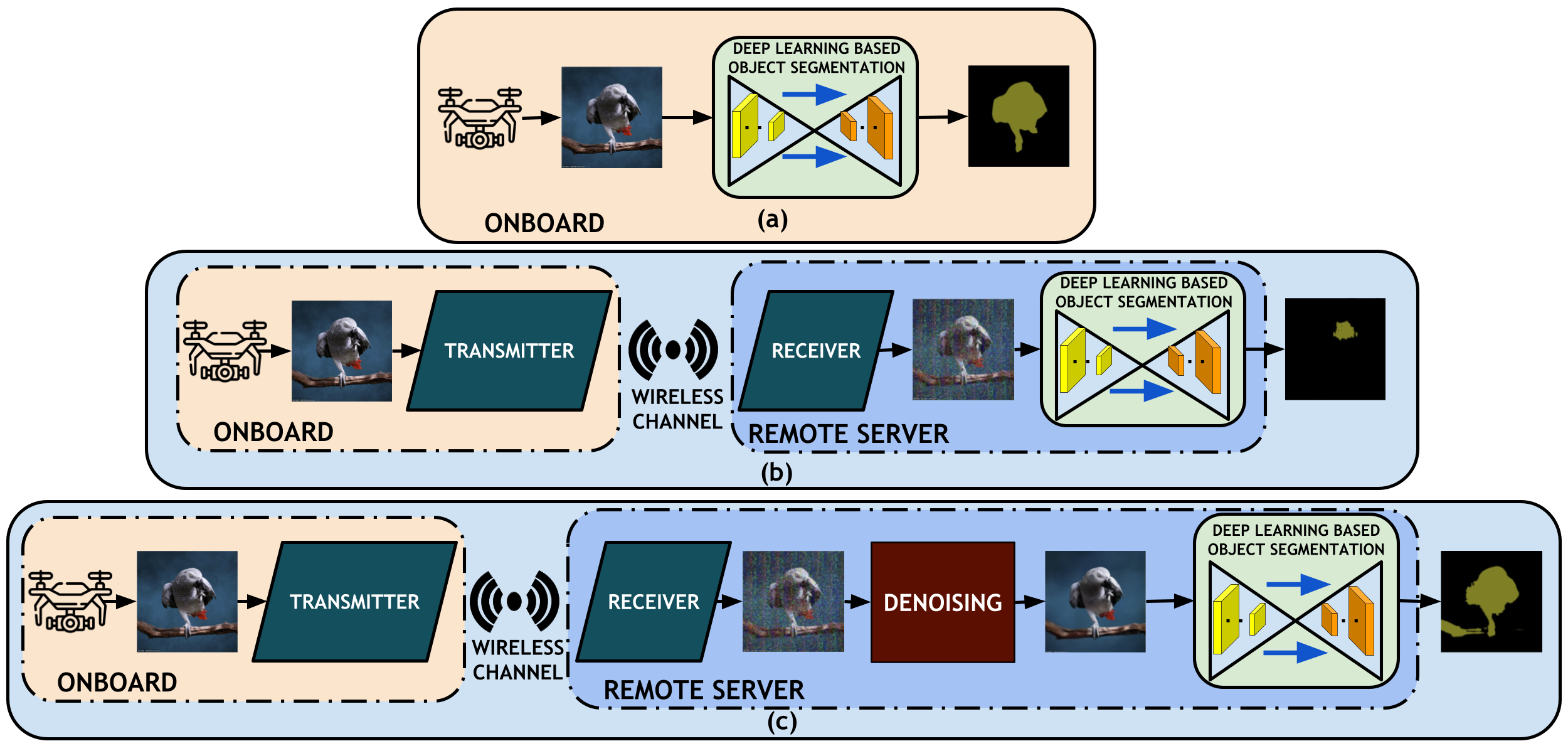}}
    \caption{\small The overview of three different scenarios for deep learning based object segmentation and detection is presented: (a) shows the case where the pretrained deep network is used directly on UAV (no offloading); (b) shows the naive offloading scheme where the received image is send to a remote server to be processed; (c) shows the proposed alternative offloading scheme in this paper. In order to eliminate the introduced noise by the communication system, a denoising stage is introduced before using the deep learning based segmentation (or detection) algorithm on the remote server.}
    \label{fig:systemcomparison}
\end{figure}

Recent developments in both hardware and software technologies made it easier to access the Unmanned Aerial Vehicles (UAVs) commercially and increased their use cases today which also increased the demand on running a large variety of algorithms on UAVs. For example, a low level computer vision algorithm can run on a camera, and an on-board computer can perform computationally heavy vision tasks (such as object detection, tracking as in \cite{albaba2020synet,sahin2021yolodrone,gozenvisual} and infrared image processing as in \cite{ozer2022siamesefuse,ozkanoglu2022infragan}) while, on the other side, a wireless communication algorithm can run to check and control more essential tasks related to the control center (such as handling the current location, current distance to the center, relaying communication messages, etc.).  However, running multiple algorithms on a UAV can require additional computational resources including dedicated additional processors (such as Graphics Processing Unit - GPU), larger memory and larger power resources. Nonetheless, due to the physical constraints, a UAV can carry up to a certain weight and consequently, its capabilities are heavily limited by its battery capacity. That makes the battery (power source) a main bottleneck for many UAV based applications. Offloading is proposed as a solution to ease such limitations by offloading different and computationally heavy tasks to a remote server. In particular, offloading deep learning based tasks are computationally more beneficial as many deep algorithms require large computational resources including processing power (GPU) and memory. Many existing state-of-the-art deep learning based algorithms are reaching a point where running such algorithms become problematic on modern desktop computers with strong GPUs, let alone running them on mobile computing devices. Optimizing deep networks for mobile devices is another active research field where the goal is reducing amount of the required computational resources for the existing state-of-the-art deep learning algorithms so that they can be used on mobile platforms (such as \cite{huang2018yolo,sandler2018mobilenetv2}). In many cases, mobile versions of such existing state-of-the-art algorithms are essentially a smaller version of the originals and they usually yield a reduced accuracy. Consequently, a trade-off between the accuracy and computational requirements must be considered in such mobile approaches. 

In many cases, multiple deep learning based algorithms are required to run simultaneously (e.g., an object detection algorithm and an object tracking algorithm can be run simultaneously while another algorithm can be run for segmentation) and that increases the need for using larger computational resources (hardware) on an UAV. In order to help with such computational burden on UAVs, offloading has been proposed as an alternative solution to help with the limited hardware and power problems of UAVs in many recent works as in \cite{chen2020intelligent,mukherjee2020distributed}. Offloading a task is the process of sending a task that was suppose to be running on UAV to a remote device (which is usually a stronger mobile device or an edge server on the ground) where the computation is done. By doing so, the goal is reducing the computational burden on the existing (on-board) computers of the UAV so that the UAV's battery can last longer, while still benefiting from the output of the offloaded task. When the offloading process is done successfully, the mobile GPU requirements can be eliminated on many UAVs allowing us to build smaller UAVs (drones). While the topic of offloading takes attention of many researchers lately, the majority of the relevant research focuses on deciding when to offload (when the load increases, etc.) from the perspective of optimizing the power usage \cite{chen2020intelligent,mukherjee2020distributed}. In such works, the assumption is that the offloaded task may not be affected by the offloading procedure. An important yet less considered problem, however, is that the performance of the offloaded task (or its output) may change based on the different aspects of the used communication system since certain tasks (such as tasks using deep learning) are more prone to noise. Deep learning based algorithms are such algorithms that are susceptible to noise as many recent papers report \cite{zha2019rank, tian2020deep}. Many UAV applications use off-the-shelf pretrained deep algorithms. Examples might include using pretrained YOLO \cite{glenn_jocher_2020_4154370} for object detection and pretrained MobileNet \cite{sandler2018mobilenetv2} for object segmentation. While such algorithms come with already trained (pretrained) models, they are trained on datasets that do not consider the noise introduced by the communication system. Therefore, it is not known how a pretrained deep algorithm (such as MobileNet) would perform against the varying noise introduced by 5G wireless systems and, consequently, the question: {\it "How reliable is your offloaded object segmentation task?"} remains unanswered.

In this paper, first, we analyze the performance of offloading deep learning based object detection and segmentation algorithms in the presence of the noise which is generated by a 5G communication system for the first time. We demonstrate that the noise introduced by the used 5G system can affect the performance of such pretrained networks significantly on the remote server side, when the images are offloaded over 5G to a remote server. We characterize the noise introduced by the used 5G wireless system with different system parameters including signal to noise ratio (SNR) and the Doppler effect (signifying the velocity of UAV). Then, we introduce possible approaches and techniques to eliminate the effect of such noise on the used pretrained networks on the remote server side. The idea is that, instead of re-training each used pretrained network (such as YOLO and MobileNet individually), we introduce an intermediate block for denoising the transmitted data (image) first and then giving that image to the already pretrained network. In that intermediate block, we first study the effect of using classical (image-processing based) filtering techniques as a remedy and then introduce a novel deep learning based denoising algorithm. Our proposed deep learning based approach: Noise-Removing Net (NR-Net) introduces the state-of-the-art denoising results to improve the performance of off-the-shelf algorithms including YOLO and MobileNet when the object detection and segmentation tasks are offloaded to a remote server over 5G. 

Fig. \ref{fig:systemcomparison} summarizes the two most commonly used scenarios for the deep learning based segmentation tasks in addition to our proposed alternative approach. In Fig. \ref{fig:systemcomparison}a, the image is directly processed (segmented) on UAV's on-board, however this  approach also drains the on-board battery faster. Fig. \ref{fig:systemcomparison}b shows an alternative approach where the image is first transmitted to a remote server over a 5G wireless communication system, and then, the received image is segmented by the pretrained deep network. Notice the effect of the noise introduced by the communication system on the output (in the figure, the offloaded task is the segmentation task and the used pretrained deep learning algorithm is: a MobileNet model \cite{sandler2018mobilenetv2}). In this paper, in addition to analyzing the affect of the noise introduced by the 5G system, we also introduce using a denoising stage to reduce or eliminate the effect of the noise introduced by the used communication system before using a pretrained (off-the-shelf) deep algorithm. For denoising the received noisy image, we propose a novel transformers based multi-stage deep architecture. Our proposed multi-stage deep architecture uses a spatial attention mechanism and yields state-of-the-art results. 

When a deep learning based task using an image as input is offloaded, the transmitted image can be used on multiple deep networks on the remote edge server. A potential alternative solution to eliminate the denoising stage is, re-training (or fine-tuning) those networks. However, the cost of re-training each such network individually is costly and time consuming. Consequently, instead of re-training each possible pretrained network with the noisy images, we introduce using only a single denoising network to eliminate the noise first and then use that resulting denoised-image on all of the existing pretrained off-the-shelf algorithms. 

Our contributions in this paper include: \textbf{(i)} studying and analyzing the effect of the noise introduced by 5G wireless systems on offloading deep learning based vision tasks (such as object segmentation and object detection) for the first time; \textbf{(ii)} introducing the use of a denoising step before applying the existing pretrained deep algorithms on the received image directly; \textbf{(iii)} introducing a novel deep architecture for denoising the received noisy image over a 5G wireless system; \textbf{(iv)} comparing NR-Net's performance to classical filtering techniques as well as existing state-of-the-art deep denoising techniques and reporting performance improvement when compared to the existing state-of-the-art algorithms.

\section{Background and Related Work}

This is one of the earliest works studying the effect of noise on offloading deep learning based object segmentation and detection tasks to a 5G edge server.  Furthermore, this is the first work that introduces a denosier block with a novel deep architecture to eliminate the effect of the channel noise introduced by the used 5G system in the presence of offloading. Therefore, relevant work is limited. The closest work is our preliminary work where the effect of offloading object segmentation task on 4G wireless systems was studied \cite{ilhan2021offloading}. There, it was reported that when offloading is done over a 4G system, the noise introduced by the 4G network can affect the performance of the used off-the-shelf deep learning based segmentation algorithm. In this work, we take that work to the next level by first studying the performance of 5G network and then by introducing a deep denoiser architecture. While 5G, practically, is the current state-of-the-art communication systems in many aspects, 6G systems are also being discussed in the literature as in \cite{kunst2020application}. However, as of today, due to the lack of practical deployment of 6G models, 6G based systems were not used in this paper. Nonetheless, the proposed framework can be easily adopted into the similar scenarios using 6G.  
\\
\textbf{OFFLOADING:} While offloading a task from UAVs has been studied heavily the literature, those papers typically considered the problems from the communication side of the systems and they mainly focused on the power management aspect of the systems as in~\cite{9622148}. There is, however, an important yet ignored aspect that should be considered when offloading a task: "{\it the effect of the noise on the offloaded task}". Therefore, we provide a short summary of the most related work from both offloading and the image denoising aspects below.

\begin{figure*}[t!]
    \centerline{\includegraphics[width=\textwidth]{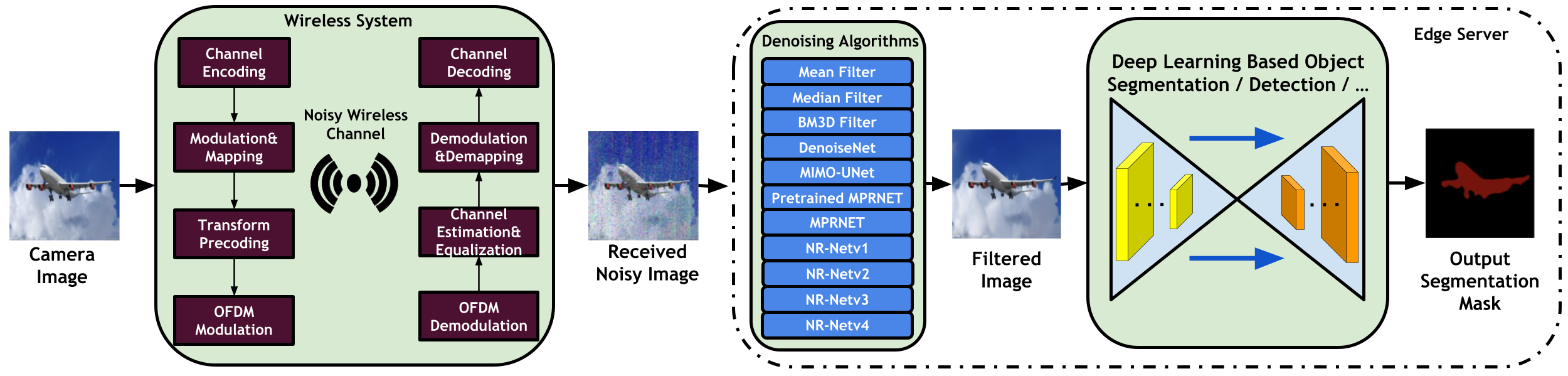}}
    \caption{System overview: the image taken by the on-board camera is transmitted to a remote server via a 5G system over a wireless channel. The received signal first goes through an image denoising stage to obtain a denoised image. The denoised image is, then, given to the offloaded deep learning task which uses a pretrained network (for example: pretrained YOLOv5 \cite{glenn_jocher_2020_4154370}, pretrained MobileNet \cite{sandler2018mobilenetv2}, etc.)}
    \label{fig:overview}
\end{figure*}

Many relevant offloading papers focus on how a UAV might be utilized as a mobile access point in an efficient way as in \cite{zeng2017energy} (mostly from the perspective of energy efficiency). As another example, the authors of ~\cite{callegaro2018optimal} focused on computing the optimal offloading conditions for different communication scenarios. The work in \cite{li2018task} introduced a method to utilize a UAV as a mobile edge server (MES) to provide task offloading services. Their main goal was looking for a way to maximize the migration throughput of the tasks. In that work, the authors introduced using a semi-Markov decision process (SMDP) without using transition probabilities. In \cite{8210823}, the hovering heights of multiple UAVs in disaster or emergency situations are studied for public safety, when the UAVs are utilized as mobile access points in 4G. Additionally, the optimal height of UAVs is studied to minimize both power and latency in \cite{costanzo2020dynamic}, when they are used as mobile edge servers. When multiple UAVs are deployed and when multiple images are needed to be taken at different locations, the work in \cite{kim2019optimal} introduced utilizing Hungarian algorithm to assign multiple tasks to those multiple UAVs. That work was also extended in \cite{kim2020machine} to assign multiple tasks to multiple UAVs, considering the situation where the total number of available UAVs is not matching to the total number of tasks. The authors proposed using K-means and reinforcement learning based techniques for the problem with keeping the energy efficiency in mind. The work in \cite{liu2020incentive} studied the problem of wireless charging of UAVs, where the microwave power transmission is utilized to charge UAVs. 

While, the image and video analysis aspect of edge computing has also been studied intensively in the literature as in ~\cite{zhang2019edge}, there is no work that focuses on studying the effect of the wireless channel parameters and of the used communication system on the performance of offloading deep learning-based tasks (such as image segmentation or detection tasks) to a remote server accept our preliminary work in \cite{ilhan2021offloading}. 
\\
\textbf{DENOISING:} Image denoising has been an active research area in the relevant deep learning based literature. Recently introduced denoising solutions evolve around deep learning based techniques. A deep convolutional neural network (DenoiseNet) was proposed in \cite{remez2017deep} to eliminate Poisson distributed noise on images. DenoiseNet consists of 20 layers including 18 ReLU (nonlinear) and 2 sigmoids (linear) layers. The output of the previous layer is convolved with 64 kernels of size 3x3 and 1 stride at each layer. The loss function of the network is mean square error and presented in Eq. \ref{eq:mse_loss}. The network is trained with Pascal VOC dataset \cite{everingham2015pascal}. Since the network works on greyscale images, colored RGB input images were converted into YCbCr. The Y channel (grey counterpart of the coloured image) is applied to the network and combined with chroma components (Cb and Cr) at the end of the network. 

MPRNET is introduced for image denoising applications including deblurring and deraining \cite{DBLP:journals/corr/abs-2102-02808}. It has a multi-stage architecture. In its first two stages, a UNet architecture is used as encoder-decoder subnetwork, and in its last stage, the Original Resolution Block (ORB) subnetwork is used. By using multiple Supervised Attention Modules (SAM) in the output of each stage, the features of the current stage are evaluated and improved with the supervision of the ground truth before they are transferred to the next stage. Therefore, MPRNET uses a progressive learning structure. Cross-Stage Feature Fusion (CSFF) mechanism helps sharing the information between the stages. Thanks to the encoder-decoder subnetworks, while the contextual information on the image is extracted, the texture information in the output image is preserved by using the ORB subnetwork, which works without changing the resolution of the image at the last stage.
The loss function of MPRNET is given in Eq. \ref{eq:mprnet_loss} for deblurring and deraining tasks whereas the denoising task solely depends on the Charbonnier loss (in our denoising experiments, we used Eq. \ref{eq:mprnet_denoising_loss} for MPRNET).

\vspace{-0.091in}
\begin{equation}
\vspace{-0.091in}
\label{eq:mprnet_loss}
\resizebox{.97\linewidth}{!}{
    $\mathcal{L_{\text{D}}} = \sum_{s=1}^{3} \{\sqrt{\lvert\lvert {\hat{I}_{s}-I_g}\rvert\rvert^{2}  + \epsilon^{2}}\} + \{\lambda \times \sqrt{\lvert\lvert {\Delta (\hat{I}_{s})- \Delta (I_g)}\rvert\rvert^{2}  + \epsilon^{2}}\}$
    }
\end{equation}
where $L_{D}$ represents the loss function used for both deblurring and deraining, $s$ is the stage index, $I_g$ is the ground-truth image, $\hat{I}_{s}$ is the predicted image at stage $s$, $\Delta$ is the Laplacian operator, $\epsilon$ and $\lambda$ are empirical constants (set to $10^{3}$ and $0.05$, respectively). 
\vspace{-0.091in}
\begin{equation}
\vspace{-0.075in}
 \mathcal{L_{\text{Denoising}}} = \sum_{s=1}^{3} \{ \sqrt{\lvert\lvert {\hat{I}_{s}-I_g}\rvert\rvert^{2}  + \epsilon^{2}} \} 
    \label{eq:mprnet_denoising_loss}
\end{equation}

\begin{figure}[!b]
    \centering
    \includegraphics[width=0.99\columnwidth]{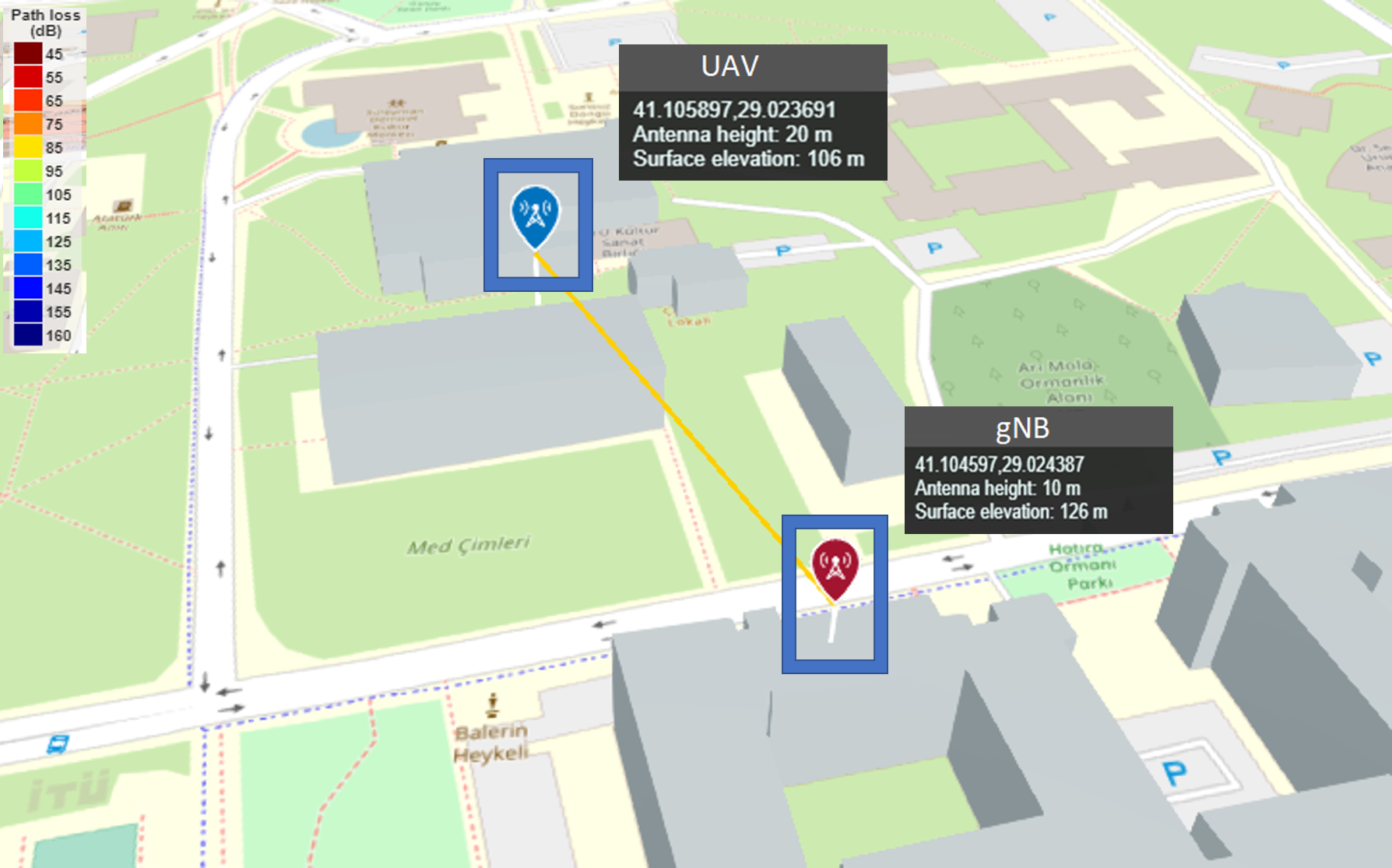}
    \caption {A set up example is shown from our 5G communication simulation which demonstrates the coordinates of the UAV and the gNB on the map. }
    \label{fig:system_map}
\end{figure}

\vspace{-0.11in}
\begin{equation}
    \vspace{-0.11in}
    \label{eq:mse_loss}
    MSE(y, \hat{y}) = {{\frac{1}{HW}\sum_{i=1}^{H}\sum_{j=1}^{W}({y_{ij} - \hat{y}_{ij}})^2}}
\end{equation}
where $y$ is the ground truth, $\hat{y}$ is the prediction (output) and $H$ and $W$ are the height and width of the ground truth, respectively. MIMO-UNet is another recent network proposed to denoise images in \cite{cho2021rethinking} based on multi-stage idea. The noisy input image goes through an encoder block which is not only downsizing the image but also extracting features by using a Shallow Convolutional Module (SCM) rather than a sub-network. Multi-input single encoder, multi-output singe decoder and Asymmetric Feature Fusion (AFF) modules are used in that network. The network is trained with the GOPRO dataset. The loss function of the network is a combination of content loss and fast Fourier transformation of the content loss which is called as Multi-Scale Frequency Reconstruction (MSFR) loss function. The loss function of the MIMO-UNet is presented in Eq. \ref{eq:mimo_loss}.
\vspace{-0.091in}
\begin{equation}
\vspace{-0.091in}
 \resizebox{.9\linewidth}{!}{
 $\mathcal{L_{\text{MIMO-UNet}}} = \frac{1}{t_k}\sum_{k=1}^K\{||\hat{S_k} - S_k||_1 + \lambda||\mathcal{F}\{\hat{S_k}\} - \mathcal{F}\{S_k\}||_1\}$
    }
    \label{eq:mimo_loss}
\end{equation}
where $t_k$ is the total elements, $K$ is the number of levels, $\hat{S_k}$ is the prediction of the $k^{th}$ level, $\mathcal{F}$ is the fast Fourier transform operation and $\lambda$ is experimentally set to 0.1.

\section{System Overview}

Fig. \ref{fig:overview} shows the overview of our proposed system where we consider and eliminate the noise introduced by the wireless communication system so that a pretrained deep algorithm based task can be offloaded efficiently. In our proposed system, first, the image is acquired by an onboard camera on UAV and then a 5G wireless communication system is used to transfer the acquired image to the remote (edge) server. On the edge server side, once received, the image goes through a denoising step and then, the resulting denoised image is used as input for the offloaded task based on a pretrained deep algorithm. Next we provide the details of our used 5G wireless communication system.

\section{5G Communication System}
\label{WirelessSection}
The evolution of mobile networks over multiple generations includes many innovations. While GSM (2G) enabled wireless voice calls, the redesigned interfaces of GPRS (2.5G), UMTS (3G) and LTE (4G) enabled wireless data connectivity and gradually improved the data rates along with the quality \cite{patel2018comparative}. 5G New Radio (NR) can be considered as the successor of the LTE wireless communication system by enabling much higher data rates and much higher efficiency for mobile broadband. It uses higher frequency radio waves such as millimeter waves (mmWaves) than previous generations to increase its bandwidth and data rate. Thanks to 5G NR, the Internet of Things (IoT) and edge computing systems will be deployed more conveniently. Three sample systems that can benefit from such high data-rates and low latency are: (1) enhanced Mobile BroadBand (eMBB) for high data rates, (2) Massive Machine Type Communications (MMTC) for the large number of connections and (3) Ultra-Reliable and Low Latency Communications (URLLC) for low latency \cite{8412469}.

\begin{figure*}[!ht]
    \centerline{\includegraphics[width=1.0\linewidth]{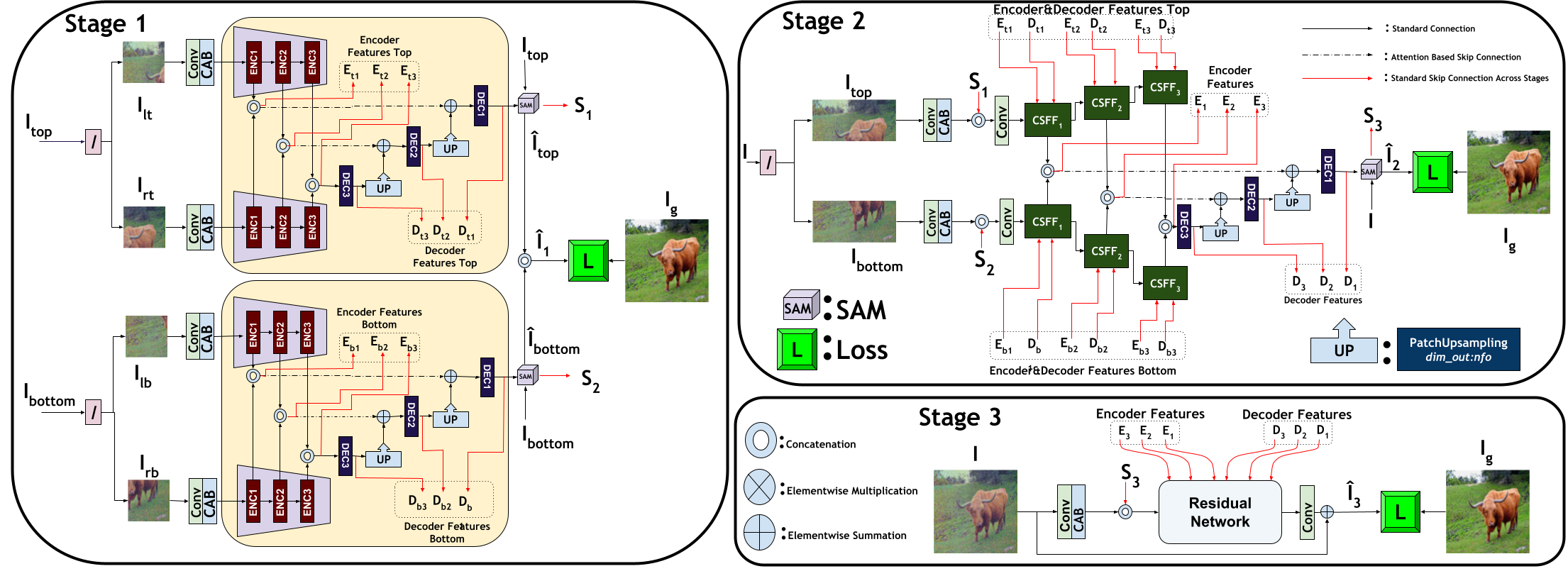}}
    \caption{The architecture of our proposed denoising algorithm: NR-Net is shown in this figure. The overall architecture consists of three interconnected stages. The first stage considers the input image as a combination of four parts, the second stage considers the input image as a combination of two parts and the third stage considers the input image as a whole input. Stage 1 shares information with Stage 2 and Stage 2 shares information with Stage 3 through the red (skip) connections. $I$ refers to the noisy images, $I_g$ is ground truth for denoised images, $\hat{I}$ is the denoised image obtained from our deep network. }
    \label{fig:networkoverview}
\end{figure*}

In this work, we study the case where computationally heavy deep learning based vision tasks are offloaded to an edge server over a 5G NR system. Our used 5G NR system's overview can be seen in Fig. \ref{fig:overview} (see the "Wireless System" block in the figure). In our system, we study a couple of scenarios where we have one UAV and one next generation NodeB (gNB) communicating over 5G NR in a rural area (see Fig. \ref{fig:system_map} for a visualization of our scenario). In that scenario, the images captured by the UAV are transmitted over Hybrid Automatic Repeat Request (HARQ) process to gNB. The HARQ process is based on the phenomenon of the stop-and-wait which means after each sent packet, an acknowledgement is waited by the transmitter before continuing with the next packet during the transmission. The used "wireless system" blocks are given in Fig. \ref{fig:overview}. There are three types of channels that we consider in 5G NR: (i) logical channels, those are responsible for data transmissions between Medium Access Control (MAC) and Radio Link Control (RLC)  layers; (ii) Transport channels, those are responsible for data transmissions between MAC and Physical (PHY) layers; and (iii) Physical channels, those are responsible to data transmissions between the different levels of the PHY layer. Since we transmitted the data from UAV to gNB, we employed the Uplink Shared Channel (UL-SCH) as transport channel and Physical Uplink Shared Channel (PUSCH) as physical channel. The UL-SCH consists of error detection and correction, rate matching, code block concatenation and Cyclic Redundancy Check (CRC). The detailed definition of each of the used blocks and further details can be found in 5G standards released by 3GPP \cite{3gpp5g}.

In our scenario, the channel encoded data by using LDPC is passing through UL-SCH and PUSCH channels respectively. LDPC is a linear error correcting code which is defined by a sparse parity check matrix \cite{gallager1962low} and UL-SCH and PUSCH channels are the transport and physical channels respectively. Then, the demodulation reference signal (DM-RS) is added to the NR grid. DM-RS is the reference signal used for channel estimation as a part of PUSCH demodulation. The NR grid is, then, passed over Multi-Input Multi-Output (MIMO) antenna system to increase the data rate. The MIMO pre-coder enables the calculation of the propagation matrix at the transmitter side by matching it to the DM-RS. After MIMO precoding, the NR grid is converted into a continuous time-domain waveform by using Cyclic Prefix Orthogonal Frequency-Division Multiplexing (CP-OFDM) modulation. In CP-OFDM, the cyclic prefix duration per symbol varies since the sub-carrier spacing varies from 15kHz to 120kHz in 5G NR.

The modulated signals are passed through a noisy CDL-A type wireless communication channel \cite{3gpp38901} since we simulate an NLOS outdoor environment in this work. On the receiver side, the signals are first demultiplexed by using CP-OFDM demodulation. Then, the channel grid is estimated and equalized by using perfect channel estimator and Minimum Mean Squared Error (MMSE) equalizer. The received NR grid is demodulated and decoded at the gNB to obtain the image data. In this paper, we assume that both gNB and the edge server are located on the same node, therefore, there is no additional communication loss between them.

\section{Image Denoising}

In our system, the received image is denoised on the edge server. Image denoising is the process of removing the noise from a noisy image \cite{fan2019brief} and, typically, spatial filters are considered as  classical solutions to denoise a given noisy image. An image denoiser can include basic techniques such as median and mean filtering, or can be more complicated such as Block-matching and 3D Filtering (BM3D) algorithm  \cite{dabov2007image}. Recent research, however, focuses on deep learning based solutions as in \cite{DBLP:journals/corr/abs-2102-02808}. In this paper, we introduce a new multi-stage transformer based denoising algorithm to obtain the state-of-the-art results.

\begin{figure*}[!ht]
    \centerline{\includegraphics[width=0.9\linewidth]{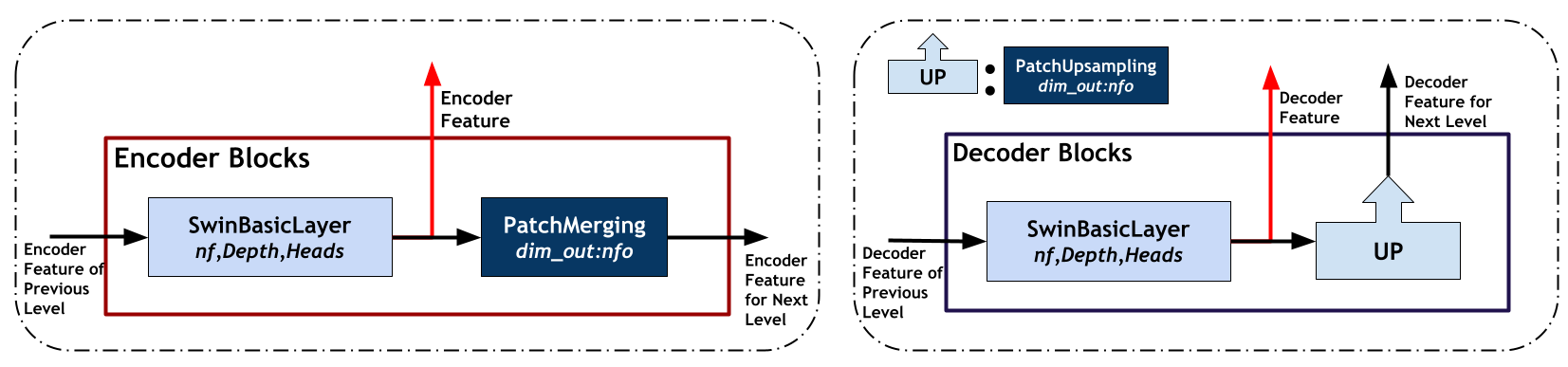}}
    \caption{The Encoder \& Decoder blocks. Left shows the encoder block details and right shows the decoder block's details.}
    \label{fig:encdecblock}
\end{figure*}

\subsection{Proposed Deep Denoiser Architecture: NR-Net}

\begin{figure*}[ht!]
    \begin{subfigure}[t!]{\linewidth}
    \centerline{\includegraphics[width=0.7\linewidth]{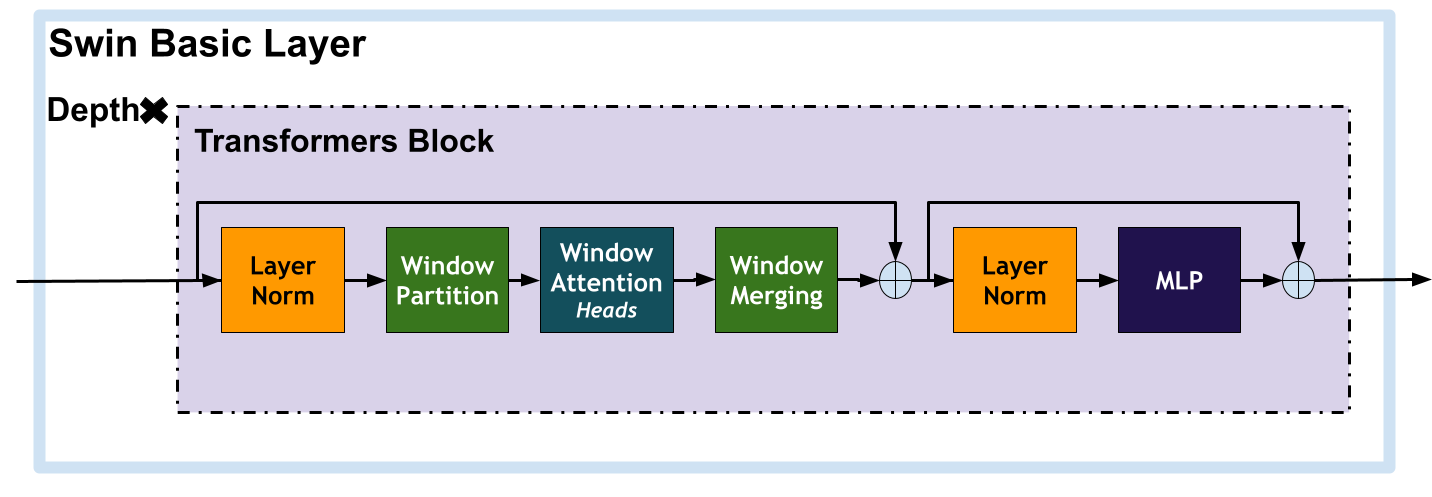}}
        \caption{The SBL architecture. In both of Fig. \ref{fig:encdecblock} and Fig. \ref{fig:csff}, the main transformer block of the SBL repeats itself '\textit{Depth}' times (see the light blue boxes in Fig. \ref{fig:encdecblock} and Fig. \ref{fig:csff}). }
        \label{fig:SwinBasicLayer}
    \end{subfigure}
    \begin{subfigure}[t!]{\linewidth}
        \centerline{\includegraphics[width=0.7\linewidth]{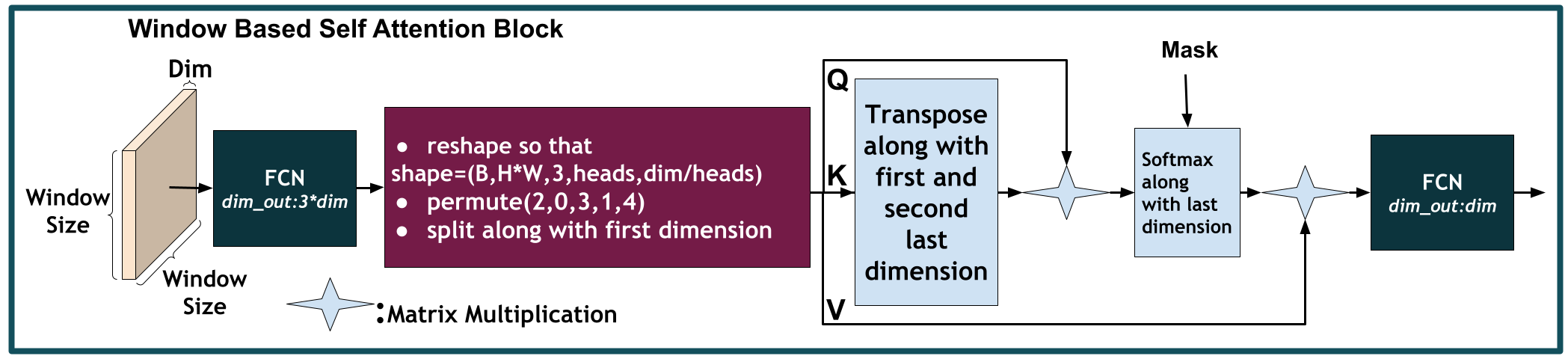}}
        \caption{Window-based Self Attention Module as used in Fig. \ref{fig:SwinBasicLayer}.}
        \label{fig:WindowAttention}
    \end{subfigure}
    \caption{(a) shows the Swin Basic Layer (SBL) module and (b) shows the Window based Self Attention module.}
\end{figure*}

We propose a transformer based multi-stage denoiser network which we name Noise-Removing Net (NR-Net) in this paper. In our network, each transformer block contains a window based multi-head self attention (W-MSA) module (see Fig. \ref{fig:WindowAttention}). NR-Net consists of three interconnected subnetworks and we name each of those subnetworks as a stage. In total, we have three stages in NR-Net, and each of those stages processes the image at a different level. For example, the stage 1 considers image parts as equally divided into four pieces; the stage 2 splits image into two halves (top half part and bottom half part) and stage 3 considers the entire image as one (whole) image. Those three stages are interconnected through various skip connections as denoted by the red arrows in Fig. \ref{fig:networkoverview}. The network is trained in an end-to-end fashion. 

The first stage (Stage 1) of the network essentially involves an Encoder-Decoder architecture (shown in yellow background). Even though there are two Encoder-Decoder architectures (i.e., two yellow boxes) in the figure, both of them are identical over which both bottom  $I_{bottom}$ and top $I_{top}$ parts of the image are passed. Similarly, each Encoder-Decoder architecture, has only one encoder (see the sequence of ENC1, ENC2 and ENC3 in the figure) and one decoder (see the sequence of DEC1, DEC2 and DEC3 in the figure). The inputs $I_{lt}$,$I_{rt}$,$I_{lb}$ and $I_{rb}$ refer to the left-top, the right-top, the left-bottom and the right-bottom part of the image, respectively. Each ENC and DEC block contains a transformer block.

\begin{figure*}[ht!]
    \begin{subfigure}[b!]{0.49\linewidth}
      \centerline{\includegraphics[width=\linewidth]{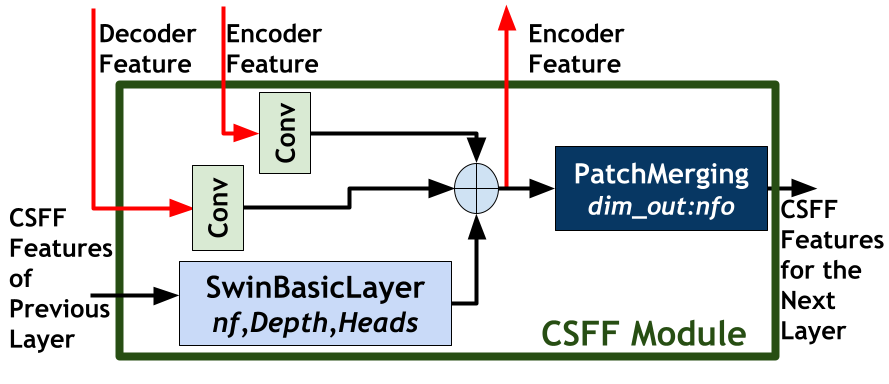}}
        \caption{CSFF module}
        \label{fig:csff}
    \end{subfigure}
    \begin{subfigure}[b!]{0.49\linewidth}
        \vspace{0.2in}
        \centerline{\includegraphics[width=\linewidth]{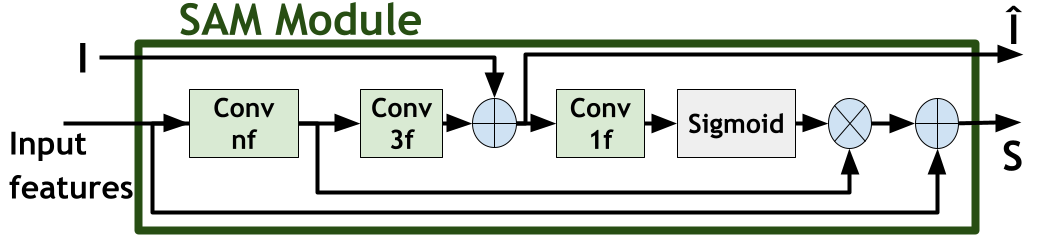}} 
        \vspace{0.15in} 
        \caption{SAM module}
        \label{fig:sam}
    \end{subfigure}
    \caption{CSFF and SAM modules. (a) Architecture of a Cross-Stage Feature Fusion (CSFF) module as used in Stage 2. (b) Architecture of the Supervised Attention Module (SAM). SAM blocks are utilized at the end of Stage 1 and Stage 2 in Fig \ref{fig:networkoverview}.}
\end{figure*}

\begin{figure*}[t!]
  \centering
  \begin{subfigure}[t]{0.48\linewidth}
   \includegraphics[width=0.9\textwidth]{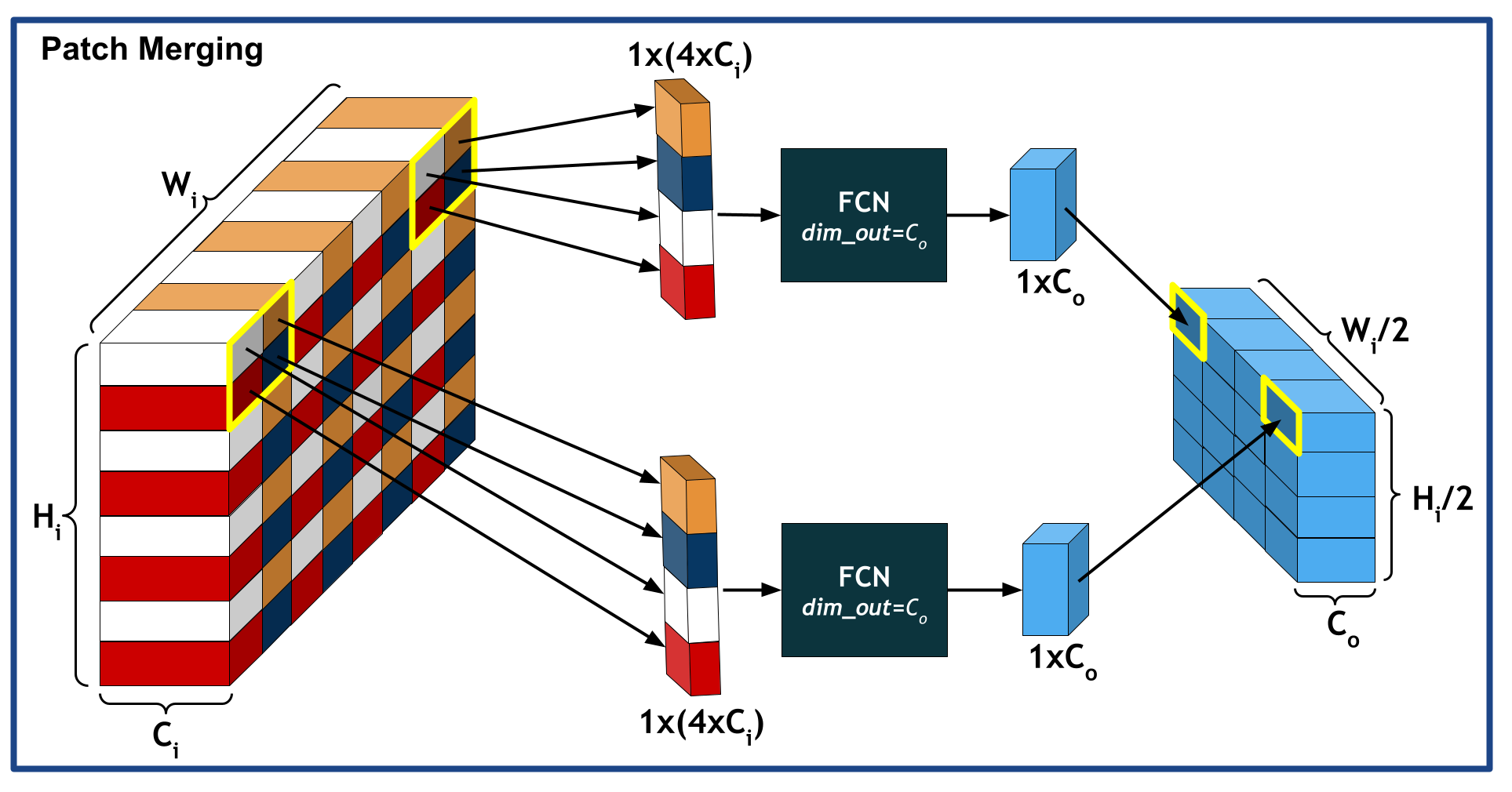} \caption{ Patch Merging} \label{fig:PatchMerge}
  \end{subfigure}
  %\hspace{0.5cm}
  \begin{subfigure}[t]{0.48\linewidth}
     \includegraphics[width=0.9\linewidth]{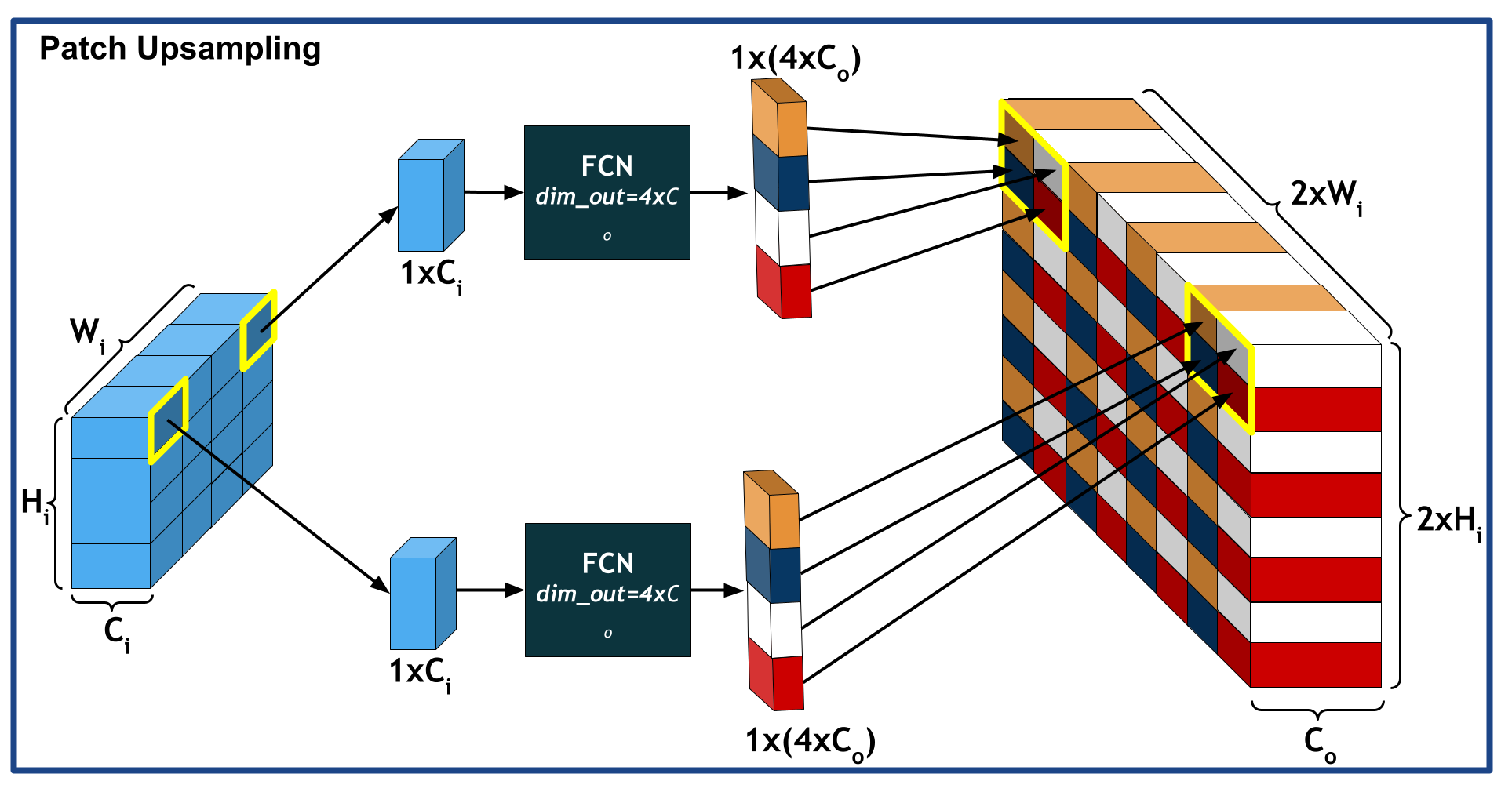} \caption{ Patch Upsampling} \label{fig:PatchUpsample}
  \end{subfigure}
  %\vspace{-0.12in}
   \caption{ (a) shows the details of Patch Upsampling Layer (PUL), (b) refers to Patch Merging Layer. $H_i$, $W_i$ are height and width of the input feature map. Besides, $C_i$ is channel size of input whereas $C_o$ denotes the channel size of output.}
     \label{fig:PatchMergeUpsample}
\end{figure*}

Stage 2 includes a Cross-Stage Feature Fusion (CSFF) block to fuse contextual information between the stages. Fig. \ref{fig:csff} shows the details of the CSFF block. Similar to ENC and DEC blocks, the CSFF block also contains a transformer block. A CSFF block first processes the input coming from Stage 1 through the transformers and then performs element-wise summation with the features ($E_b$ and $E_t$) of the encoder of Stage 1. The result of the summation operation is combined with the result of the other branch and then shared with Stage 3 (see $E_1$,$E_2$,$E_3$ in Stage 2 and in Stage 3 of Fig. \ref{fig:networkoverview}). At the same time, the output of the summation operation also goes through Patch Merging for downsampling. Both stages 1 and 2 have similar (but not identical) decoder blocks. Each decoder block (DEC1, DEC2 and DEC3) consists of a transformer block (See Fig. \ref{fig:SwinBasicLayer}) and a Patch Upsampling Layer (PUL) (See Fig. \ref{fig:PatchUpsample}). PUL upscales the input and for each single input pixel, it produces a set of four neighboring pixels via Fully Connected Network (FCN), therefore, can be considered as a linear interpolation operation. Consequently, the spatial dimension of resulting image becomes two times of the input image's dimension.
While PUL upscales the input, in our network, to downscale the input, we use a Patch Merging Layer in our encoder blocks (ENC1, ENC2, ENC3) and in the CSFF blocks (See Fig. \ref{fig:csff}). Fig. \ref{fig:PatchMerge} shows the architecture of a Patch Merging Layer (PML). For a given image, PML first takes four neighbouring pixels and concatenates them into a vector, and then, it feeds that vector into a FCN layer to produce a single (summary) pixel. That operation is equivalent to using a convolutional layer with 2x2 kernel size with stride 2. Furthermore, both Stage 1 and Stage 2 use Supervised Attention Module (SAM) to output stage predictions: $\hat{I}_1$, $\hat{I}_2$ and stage features which are named as $S_1$, $S_2$, $S_3$ in Fig. \ref{fig:networkoverview}. Fig. \ref{fig:sam} denotes the architecture of a SAM block. The block has two inputs: I and feature maps where I represents the noisy input image. The input feature maps go through two convolutional layers to obtain the noise information and then combined with I to obtain the predicted output: $\hat{I}$. Then, $\hat{I}$ is given as input to the next convolutional layer which also uses the output of the sigmoid function as attention heatmap. Those attention-based features are added to the input features to obtain the stage features $S$ which is shared with the next stage.

Stage 3 is the last stage in our network. It considers the whole input image without splitting it into subparts and optimizes the network by calculating the loss between the denoised image and the ground truth image where the denoised image is denoted by $\hat{I}_1$, $\hat{I}_2$, $\hat{I}_3$ where the index refers to the stage number in Fig. \ref{fig:networkoverview}.\\
\textbf{Transformer block:} Channel Attention Layer (CAL) is an attention block that learns paying attention (focus) to particular channels rather than paying attention to a particular location on each channel. Therefore,  it is not designed to focus on the spatial relationships between the pixels. However, spatial relation is informative and important in many image processing applications. Typically, in previous similar multi-stage architectures (such as MPRNET), a CAL based encoder and decoder architecture is used in each stage. Therefore, here, we replace those CAL based encoder and decoder architectures with spatially aware (visual) transformers. CAL learns to map each 2D channel information into a single scalar value along each channel by using Global Average Pooling (GAP). Then, the learned channel attentions (as a vector) are multiplied with the input as shown below: 
\begin{equation}
    \begin{aligned}
        W_A = & GAP(X), \text{ and } CAL(X)_{h_i,w_i} = & W_A * X_{h_i,w_i} \\
    \end{aligned}
    \label{eq:cab}
\end{equation}
where $X \in \mathbb{R}^{C*H*W}$ is the input and $W_A = GAP(X) \in \mathbb{R}^{C*1*1}$ is the channel attention weight. Furthermore, $h_i,w_i$ are the height and width indices for the input X, respectively.\\
In order to include better spatial capability, a new architecture based on Swin Basic Layer (SBL) \cite{liu2021Swin} was designed in this paper. SBL is used in the ENC1, ENC2, ENC3, DEC1, DEC2, DEC3 (see Fig. \ref{fig:networkoverview}) and in the CSFF blocks (see Fig. \ref{fig:csff}).  
Fig. \ref{fig:csff} shows the architecture of our CSFF block which is used in Stage 2 (the dark green blocks in Fig. \ref{fig:overview}) to share the contextual information between all stages. It takes Encoder and Decoder features obtained from Stage 1 and passes the processed Encoder and Decoder features into the residual networks in Stage 3. It first uses a Swin Basic Layer to visually process the information obtained at the current stage (Stage 2) along with the information coming from previous stage and then it passes that fused information to the next stage (Stage 3). 
Fig. \ref{fig:SwinBasicLayer} shows the architecture of a Swin Basic Layer \cite{liu2021Swin} which uses multiple transformer blocks based on the hyperparameter: \textit{depth}. These transformer blocks contain post attention layer normalization \cite{layernorm}, Window Partition function \cite{liu2021Swin}, Window Attention blocks (see Fig. \ref{fig:WindowAttention}) and a multi-layer perceptron (MLP) network. 

As opposed to using CAL, we propose to include a window based self attention module to consider the input feature spatially where the window attention weight $W_{A_w} \in  \mathbb{R}^{Heads*(H_w*W_w)*(H_w*W_w)}$ where $H_w, W_w$ are the height and the width of the predefined window, respectively. $Heads$ is a hyperparameter which represents the number of the repetitions of using window based self attention as in \cite{attentionisallyouneed}. Since that approach considers only the window as opposed to considering the entire image, the window based approach helps to avoid time complexity of the whole attention module while utilizing the benefit of the Multi Self Attention (MSA) Module. In contrast to MSA, Window based Self Attention Module (W-MSA) splits the input into a number of equal windows (patches). W-MSA (see Fig. \ref{fig:WindowAttention}) is defined as follows:
\vspace{-0.091in}
\begin{equation}
    \begin{aligned}
    \{Q,K,V\} = & FCN_1(X_w) \\
    W_{A_w} = & softmax(\frac{Q \cdot K^T}{\sqrt{h\_dim}})\\
    Y =& W_{A_w} \cdot V \\
    Y^\prime =& Y.tranpose() & \text{\tiny along with 1st and 2nd dimension} \\
    Y^{\prime\prime} = & Y^\prime.flatten() & \text{\tiny along with 1st and 2nd dimension} \\ 
    WMSA(X) =& FCN_2 (Y^{\prime\prime})
    \end{aligned}
    \label{eq:windowattention}
\end{equation}
where $FCN_1$ and $FCN_2$ are fully connected layers used in the W-MSA module. Besides, $X_w \in \mathbb{R}^{(H_w*W_w)*dim} $ is the input window, $Q, K, V \in \mathbb{R}^{heads*(H_w*W_w)*h\_dim}$. Moreover, $h\_dim=C/heads$ refers to channel size of individual head. Similarly, $Y \in \mathbb{R}^{heads*(H_w*W_w)*h\_dim}$, is first converted into $Y^\prime \in \mathbb{R}^{(H_w*W_w)*heads*h\_dim}$ and then into $ Y^{\prime\prime} \in \mathbb{R}^{(H_w*W_w)*(heads*h\_dim)}$ to combine different heads. W-MSA is placed between window partition and window merging function in Fig. \ref{fig:SwinBasicLayer}. Window partition block converts an entire input matrix into multiple window based matrices to be processed individually; and the window merging block constructs back the the entire input matrix by using all the window matrices.

\renewcommand{\arraystretch}{2}											
\begin{table}[b!]								
\begin{center}											
\caption{Comparison of the components of various networks. $SBL_{Enc}$ refers to Swin Basic Layer in the encoder, while  $SBL_{Dec}$ is in the decoder. WS is the window size in the window based multi-head self attention (W-MSA). The * means that the depth of SBL is increased when compared to NR-Netv1, NR-Netv2 and NR-Netv3.} 
\label{table:netw_comparision}
\resizebox{\columnwidth}{!}{ 											
\begin{tabular}{ |c|c|c|c|c|c|c|c|c|c|c|c|c| }
\hline											
$Networks$ 	 &$UNet$ 	 &$SAM$  	 &$SCM$ 	 &$CSFF$ 	 &$PML$ 	&$PUL$	&$AFF$ 	&$SBL_{Enc}$	&$SBL_{Dec}$	&$CAL$ 	 &$WS$ &$Loss$\\ \hline
$DenoiseNet$ 	  &	  &	  &	  &	  &	  &	  &	  &	  &	  &	 &N/A&$\mathcal{L}_{MSE}$ \\ \hline
$MIMO-UNet$  	&\checkmark 	&	& \checkmark 	&	&	&	&\checkmark	&	&	&	 &N/A& $\mathcal{L}_{content}$ + $\mathcal{L}_{MSFR}$ \\ \hline
$MPRNET$     	&\checkmark 	 &\checkmark 	  &	& \checkmark 	&	&	&	&	&	&\checkmark	 &N/A& $\mathcal{L}_{charbonnier}$ \\ \hline
$NR-Netv1$   	&\checkmark 	 &\checkmark 	  &	 &\checkmark 	&\checkmark	&	&	&\checkmark	&	&\checkmark	 &7& $\mathcal{L}_{charbonnier}$ \\ \hline
$NR-Netv2$   	&\checkmark 	 &\checkmark 	  &	 &\checkmark 	&\checkmark	&\checkmark	&	&\checkmark	&\checkmark	&\checkmark	  &7&$\mathcal{L}_{charbonnier}$ \\ \hline
$NR-Netv3$   	&\checkmark 	 &\checkmark 	&	 &\checkmark 	&\checkmark	&\checkmark	&	&\checkmark	&\checkmark	&\checkmark	  &7&$\mathcal{L}_{charbonnier}$ + $\mathcal{L}_{SSIM} $\\ \hline
$NR-Netv4$   	&\checkmark 	 &\checkmark 	  &	 &\checkmark 	&\checkmark	&\checkmark	&	&\checkmark	* &\checkmark *	&\checkmark	  &11& $\mathcal{L}_{charbonnier}$ \\ \hline
\end{tabular}\label{tab:netwcomp}						
}											
\end{center}
\end{table}											
\renewcommand{\arraystretch}{1}								

In total, we use four different versions of our noise-removing network (as an ablation study). Table \ref{tab:netwcomp} shows details of those different versions. As seen in the table, NR-Netv1 utilizes only SBL in the encoder part of the first two stages. In addition to that, NR-Netv2 also includes SBL in the decoder part of the first two stages as in the encoder. In addition to second version, NR-Netv3 uses SSIM ($\mathcal{L}_{SSIM}$) as additional loss term over $\mathcal{L}_{charbonnier}$ (See Eq. \ref{eq:mprnetssimloss} for further details). NR-Netv4 is a deeper version of NR-Netv2 because of utilizing more transformers blocks in SBLs and expanding window size to 11 in W-MSA modules. 
\begin{equation}
    \mathcal{L}_{SSIM} = 1 - SSIM(I_g, \hat{I}) 
    \label{eq:ssimloss}
\end{equation} 
where $SSIM(I_g, \hat{I})$ defined in Eq. \ref{eq:ssim}. The final loss function of the MPRNET-SSIM becomes:

\begin{equation}
    \mathcal{L}_{MPRNET+SSIM} = \mathcal{L_{\text{Denoising}}} +  \lambda_{S}*\mathcal{L}_{SSIM}
    \label{eq:mprnetssimloss}
\end{equation}
where $\lambda_{S}$ is the coefficient for $\mathcal{L}_{SSIM}$ term, ( $\lambda_{S}$ value was set to $1e^{-3}$ in our experiments).

\section{Experiments}
Our experiments are designed to analyze and eliminate the effect of the noise caused by the used 5G communication system. In this section, first, the performance metrics and the datasets as used in our experiments are defined. Then, (i) we analyze the effect of the noise introduced by the used wireless communication system, when a pretrained off-the-shelf deep algorithm is used; (ii) after that, we study how using various denoising approaches affect on the results. We conduct our experiments on two different offloaded tasks: object segmentation and  object detection. For object segmentation we choose pretrained model of MobileNetv2 \cite{sandler2018mobilenetv2} and for object detection we use pretrained model of YOLOv5 \cite{glenn_jocher_2020_4154370}. We study the performance of NR-Net in four different configurations, namely: NR-Netv1, NR-Netv2, NR-Netv3 and NR-Netv4. The differences between those configurations are given in Table \ref{table:netw_comparision}.

\subsection{5G Wireless System Parameters in Our Experiments}

Our simulated 5G NR network (as described in Section \ref{WirelessSection}) is created by using the Matlab/5G NR toolbox. Different noise levels and UAV speeds are chosen to run the simulation at different parameters (see Table \ref{table:parameters}). An image received by the UAV is transferred to the gNB using 5G PHY Uplink over a fading wireless communication channel. Since a standard 5G network is shared by multiple devices such as user equipment (UE) and IoT equipment, we allocated reasonable bandwidth (10MHz) using 52 PRBs and 15 kHz SubCarrier Spacing in our 5G simulation. Target Code Rate 600/1024 and 64 QAM modulation are used to reduce the noise sensitivity and to minimize the time needed for the UAV to hover for offloading (without the need to land for offloading). Other parameters of the used 5G system are also listed in Table \ref{table:parameters}. Clustered Delay Line - A (CDL-A) is chosen as the wireless communication channel type and Additive White Gaussian Noise (AWGN) at various Signal to Noise Ratio (SNR) levels (varying between  1dB and 20dB) is added to the channel. Since we simulate the scenario where the UAV is in motion while offloading the deep learning task, the transmitted images are affected by the Doppler Shift over 5G wireless system. Varying Doppler shift values from 100Hz to 750Hz are added to the transmitted data to examine the effect of the UAV's speed on the segmentation and detection results on the edge server.

\begin{table}[b!]
    \centering
    \caption{Used simulation parameters in our experiments.}
    \resizebox{\columnwidth}{!}{ 
    \begin{tabular}{ll} \hline
        Code Rate & 600/1024\\ \hline
        Cyclic Prefix & Normal\\ \hline
        Number of Transmit Antenna & 1\\ \hline
        Number of Receive Antenna & 2\\ \hline
        PUSCH Mapping Type & Type A \\ \hline
        Duplex Mode & FDD\\ \hline
        Resource Block & 52\\ \hline
        Subcarrier Spacing & 15 KHz\\ \hline
        Band Width & 10 MHz\\ \hline
        Modulation & 64 QAM \\ \hline
        Channel Type & Clustered Delay Line \\ \hline
        Channel Delay Spread & 30$e^{-9}$ s \\ \hline
        SNR (dB) & {1, 2, 3, 4, 5, 6 , 10, 15, 18, 20} \\ \hline
        Doppler Shift (Hz) & {100, 300, 350, 400, 500, 750} \\ \hline
    \end{tabular}
    }
    \label{table:parameters}
\end{table}

\subsection{Used Metrics And Dataset}

\begin{figure*}
    \centering
    \includegraphics[width=0.9\textwidth]{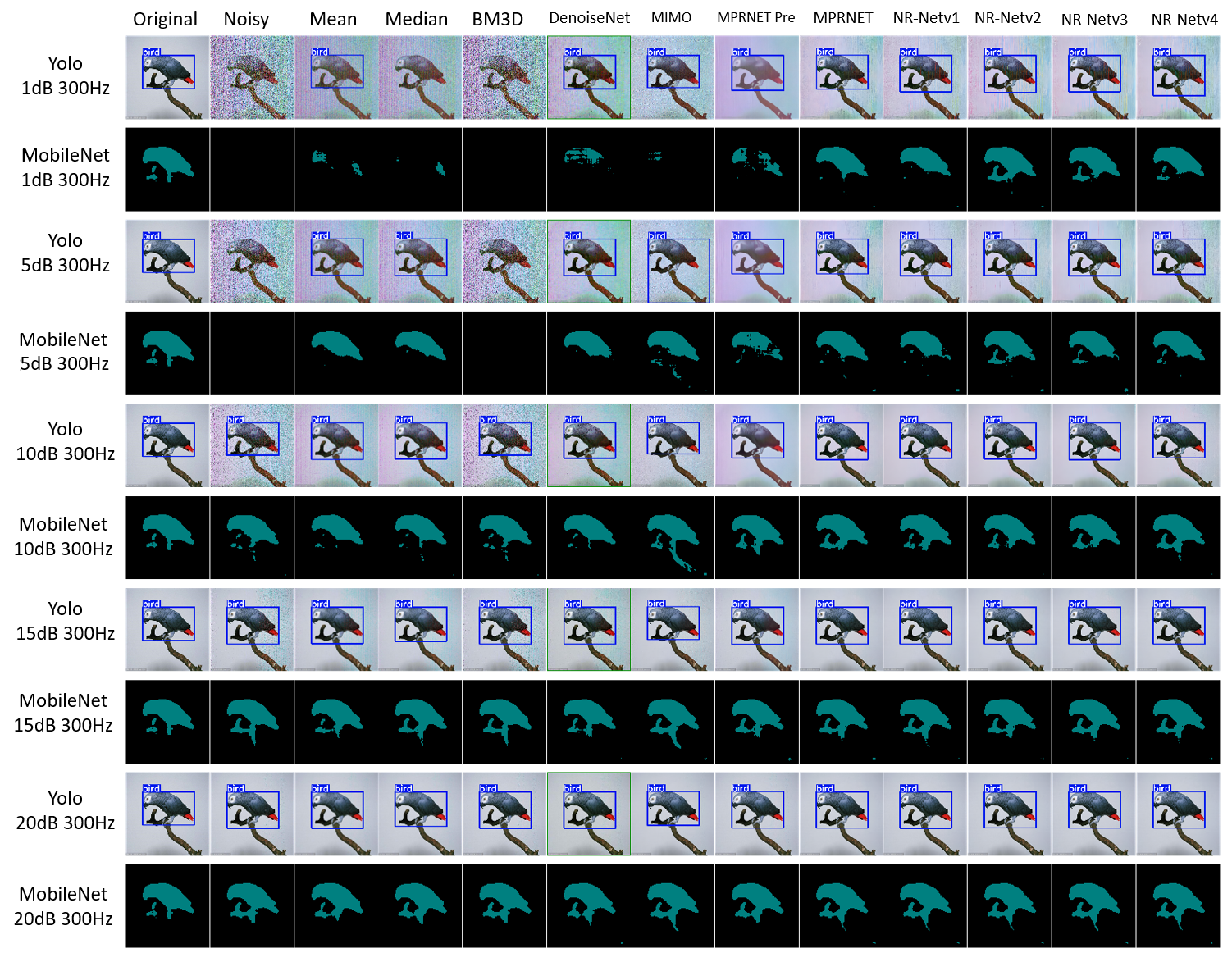}
    \caption {Comparison of qualitative results of pretrained YOLO and MobileNet algorithms on a received sample image at different SNR values. The first column shows the original image (without any noise) with the detection ground truth. The second column shows the noisy image obtained at the receiver side (before the denoiser stage). The third, fourth and fifth columns show the results obtained by using the classical denoising algorithms including mean, median and BM3D filtering in the denoising block, respectively. The images shown in the 6th, 7th, 8th and 9th columns represent the outputs obtained after using recent deep learning based denoiser algorithms in the denoiser block including DenoiseNet \cite{remez2017deep}, MIMO-UNet \cite{cho2021rethinking} and MPRNET \cite{DBLP:journals/corr/abs-2102-02808}, respectively. The remaining columns show the results obtained by using different versions of NR-Net in the denoiser block. Second rows show the segmentation results for each image shown in each column. The blue bounding box in each image represents the detection result of the YOLO algorithm on that particular image (missing bounding box means no detection).}
    \label{fig:result}
\end{figure*}

In this paper, two main deep learning based tasks were used to offload, namely object detection and object segmentation tasks. To assess the performance of our proposed denoising algorithm for object segmentation, we use multiple metrics. Those metrics are: (I) Intersection over Union (IoU) which computes the similarity between the algorithm's prediction vs. ground truth masks; (II) Structural Similarity Index Measure (SSIM) which compares the original (camera originated) image's structural information to the one which is used as the input to the pretrained deep network on the edge server, (III) Peak-Signal-to-Noise Ratio (PSNR), and (IV) mean Average Precision (mAP) which is a measure of how well the predicted bounding boxes match the ground truth on average.  

We use \textbf{FSS-1000} dataset in our experiments. \textbf{FSS-1000} dataset is developed for few-shot image segmentation applications \cite{li2020fss}. The dataset consists of 1000 classes varying from tiny objects to wild animals. Each class contains ten image and annotation pairs. Each image has the resolution of 224x224 pixels. We used 16 randomly selected classes from the dataset and generated their multiple noisy versions by transmitting the original images at different system parameters. The generated noisy images represent the received images on the receiver side (at the edge server) in our simulated wireless 5G communication system. We used 10 levels of SNR and 6 levels of Doppler Shifts in our simulations, thus, each image yielded 60 noisy versions. At the end, we obtained 18919 images for training and 57239 images for testing. We chose those numbers based on the following: in our preliminary experiments, we observed that (as our results suggest) around 18919 images was sufficient to obtain satisfactory results, therefore we aimed to use as many images as we can to obtain more representative and statistically covering test results. That idea yielded us 57239 images for testing.

To study the effect of the noise of the used 5G wireless communication system on the pretrained deep networks, we used two well known architectures: MobileNetv2 \cite{sandler2018mobilenetv2} and YOLOv5 \cite{glenn_jocher_2020_4154370}. Both architectures have pretrained models. MobileNetv2 is trained on Pascal VOC dataset \cite{everingham2015pascal} and YOLOv5 is trained on MS COCO dataset \cite{lin2015microsoft}. 

Next, we briefly define the used metrics in our experiments.
\textbf{PSNR} is a metric for measuring the ratio of the pixel values coming from two images. It is measured by the logarithm of the Mean Square Error (MSE) value of the variation between the pixels. The higher the PSNR value, the smaller the variation between the images. When one of these images is given as a reference, the noise ratio on the other image is measured \cite{5596999}. PSNR is defined in Eq. \ref{eq:psnr}.
\vspace{-0.091in}
\begin{equation}
\vspace{-0.091in}
    \label{eq:psnr}
    PSNR(y, \hat{y}) = 10\log_{10}(\frac{R^{2}} {MSE(y, \hat{y})})
\end{equation}
where $y$ is the ground truth (original image taken by the camera), $\hat{y}$ is the prediction (output image of the denoising block) and $R$ is the maximum intensity value.

\textbf{IoU} (see Eq. \ref{eq:iou}) is the metric for measuring how well a predicted area matches to the ground truth on an image. It can be used to measure how well the segmented area matches to the ground truth.  It is the ratio of the intersection and union areas between the ground truth and the prediction. The mean IoU (mIoU) is the average value of the IoU values over all of the results.
\vspace{-0.091in}
\begin{equation}
%\vspace{-0.08in}
    \label{eq:iou}
    IoU = \frac{Ground Truth Area \cap Predicted Area}{Ground Truth Area\cup Predicted Area }
\end{equation}
\textbf{SSIM} is another metric used to measure the similarity between two segmentation masks. SSIM aims to compute the structural similarity between the images by using the luminance and contrast masks, which are closer to the human visual system \cite{1284395}. SSIM is defined in Eq. \ref{eq:ssim}.
%\vspace{-0.091in}
\begin{equation}
%\vspace{-0.091in}
    \label{eq:ssim}
    SSIM(y, \hat{y}) = \frac{(2\mu_y\mu_{\hat{y}} + C_1) (2\sigma_{y\hat{y}}+C_2)} {(\mu_y^2+ \mu_{\hat{y}}^2 + C_1)(\sigma_y^2 + \sigma^2_{\hat{y}} + C_2)}
\end{equation}
where $y$ is ground truth, $\hat{y}$ is the prediction. $\mu_y$ and $\mu_{\hat{y}}$ are average intensity of ground truth and prediction, respectively. Similarly, $\sigma_y$ and $\sigma_{\hat{y}}$ are the standard deviation of ground truth and prediction, respectively. $C_1$ and $C_2$ are constants.

\textbf{mAP} is a common metric used to measure the performance of object detection algorithms. IoU is a measure assessing how well ground-truth bounding box overlaps with the predicted bounding box of an object. By using the IoU threshold true positive, false positive and false negative predictions are determined and after that, the precision and recall metrics are calculated. The Average Precision (AP) is the area under the curve of precision-recall plot and mAP is the mean Average Precision value \cite{10.1007/978-3-642-40994-3_29}.

\subsection{Results}
In this subsection, we first analyze the effect of the signal to noise ratio and the Doppler shift over our simulated wireless communication channel and we observe that the performance of the pretrained off-the-shelf deep network varies with respect to those values. Then, we introduce a denoiser block at the edge server to increase the performance of the offloaded task. In the denoiser block, we use and study the performance of multiple denoising algorithms varying from the basics to the recently proposed deep-learning based solutions. Fig. \ref{fig:result} summarizes our qualitative results while Table \ref{table:snr_segmentation_results}, \ref{table:doppler_segmentation_results}, 
\ref{table:snr_detection_results} and \ref{table:doppler_detection_results} summarize our quantitative results.

\subsubsection{Analyzing the effect of the noise introduced during offloading the deep learning based tasks}

\begin{figure*}[t!]
    \centering
    \begin{subfigure}[t]{0.246\textwidth}
        \centering
        \includegraphics[width=\textwidth]{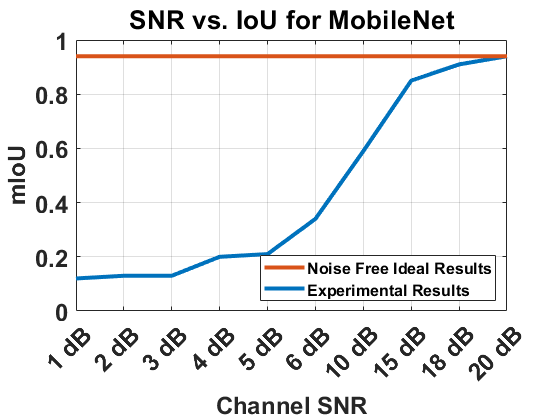}
        \caption {SNR vs. IoU}
        \label{fig:iou_snr}
    \end{subfigure}
    \hfill
    \centering
    \begin{subfigure}[t]{0.245\textwidth}
        \centering
        \includegraphics[width=\textwidth]{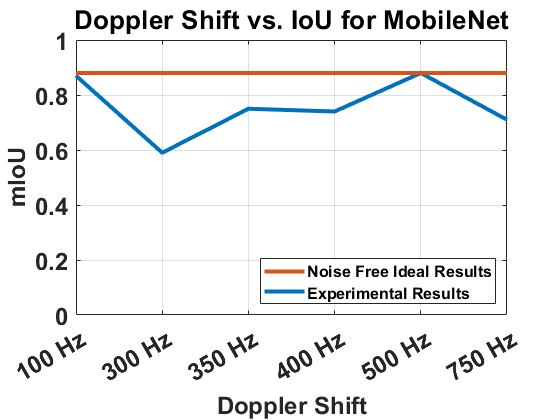}
        \caption {Doppler Shift vs. IoU}
        \label{fig:iou_doppler}
    \end{subfigure}
    \hfill
    \centering
    \begin{subfigure}[t]{0.246\textwidth}
        \centering
        \includegraphics[width=\textwidth]{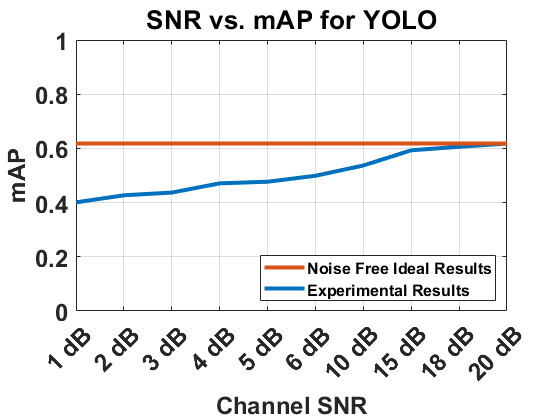}
        \caption{SNR vs.mAP}
        \label{fig:map_snr}
    \end{subfigure}
    \hfill
    \begin{subfigure}[t]{0.245\textwidth}
        \centering
        \includegraphics[width=\textwidth]{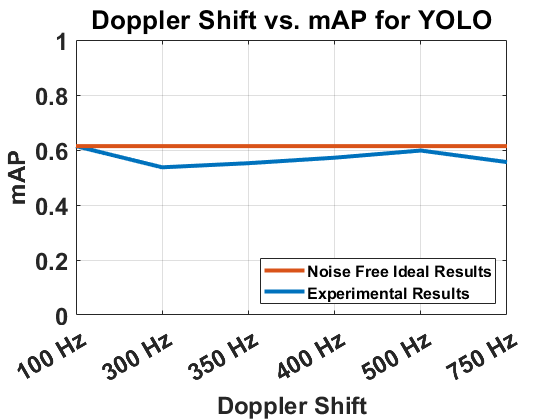}
        \caption{Doppler Shift vs. mAP}
        \label{fig:map_doppler}
    \end{subfigure}
    \caption{Performance analysis of pretrained MobileNet and YOLOv5 at different system parameters (without using any denoiser) is shown as plots. (a) shows the performance of pretrained  MobileNetv2 (mIoU) vs. SNR; (b) shows the performance of pretrained  MobileNetv2 (mIoU) vs. Doppler shift value; (c) shows the pretrained YOLOv5 performance (mAP) vs. SNR; and similarly, (d) shows the pretrained YOLOv5 performance (mAP) vs. Doppler shift value. The red line indicates the best result and is used for comparison. Blue line indicates the experimental results obtained at different values in each plot.}
    \label{fig:detection_res}
\end{figure*}

\renewcommand{\arraystretch}{1}
  \begin{table*}[t!] 														 							
      \centering 														 								
      \caption{Results of \textbf{object segmentation} using \textbf{MobileNetv2} with respect to varying channel noise SNR} 									
      \scalebox{0.8}{    	% |c|c|c|c|c|c|c|c|c|c|c|c|c|c|c|													 						
          \begin{tabular}{ |l|l|l|l|l|l|l|l|l|l|l|l|l|l|l|} 											
              \hline
              SNR& &  No Filter&  Mean&  Median&  BM3D&  DenoiseNet&MIMO-UNet&MPRNET Pre&MPRNET& NR-Netv1&NR-Netv2&NR-Netv3&NR-Netv4\\ 
              \hline
              \multirow{3}{2em}{1dB} & PSNR &13.53&17.31&16.44&16.44&17.18&19.78&15.38&22.71& \textbf{22.78} &22.55&22.37&22.18\\ 
              & SSIM &0.19&0.61&0.64&0.56&0.66&0.56&0.74&0.78&0.78&0.77&0.77&\textbf{0.80}\\ 
              & IoU &0.12&0.19&0.20&0.22&0.33&0.31&0.32&0.46&0.48&0.43&0.50&\textbf{0.52}\\ 
              \hline
              \multirow{3}{2em}{2dB} & PSNR &14.02&17.90&17.01&17.01&17.36&19.96&15.92&23.21&\textbf{23.51}&23.19&23.04&22.44\\ 
              & SSIM &0.21&0.64&0.64&0.57&0.69&0.60&0.75&0.80&0.80&0.79&0.79&\textbf{0.82}\\ 
              & IoU &0.13&0.18&0.26&0.21&0.38&0.35&0.35&0.50&0.47&0.47&0.47&\textbf{0.52}\\ 
              \hline
              \multirow{3}{2em}{3dB} & PSNR &14.59&18.52&17.61&17.61&17.51&21.06&16.52&23.65&\textbf{24.17}&23.79&23.69&23.71\\ 
              & SSIM &0.23&0.66&0.67&0.59&0.71&0.61&0.77&0.82&0.82&0.81&0.81&\textbf{0.83}\\ 
              & IoU &0.13&0.24&0.30&0.24&0.44&0.40&0.38&0.55&0.54&0.53&0.54&\textbf{0.64}\\ 
              \hline
              \multirow{3}{2em}{4dB} & PSNR &15.21&19.18&18.24&18.24&17.67&21.45&17.14&24.29&\textbf{25.02}&24.61&24.48&24.28\\ 
              & SSIM &0.26&0.69&0.70&0.61&0.73&0.64&0.78&0.83&\textbf{0.85}&0.83&0.84&\textbf{0.85}\\ 
              & IoU &0.20&0.28&0.34&0.28&0.48&0.45&0.41&0.62&0.62&0.59&0.59&\textbf{0.64}\\ 
              \hline
              \multirow{3}{2em}{5dB} & PSNR &15.87&19.86&18.90&18.91&17.87&21.87&17.79&25.29& \textbf{26.08}&25.68&25.45&25.06\\ 
              & SSIM &0.29&0.72&0.72&0.63&0.75&0.68&0.80&0.86& \textbf{0.87}&0.86&0.86&\textbf{0.87}\\ 
              & IoU &0.21&0.35&0.42&0.32&0.50&0.47&0.48&0.70&0.66&0.68&0.66&\textbf{0.75}\\ 
              \hline
              \multirow{3}{2em}{6dB} & PSNR &16.57&20.56&19.60&19.61&18.09&22.51&18.46&26.64& \textbf{27.37}&26.96&26.60&26.23\\ 
              & SSIM &0.32&0.74&0.75&0.66&0.77&0.71&0.81&0.88& \textbf{0.90}&0.89&0.89&0.87\\ 
              & IoU &0.34&0.39&0.49&0.43&0.54&0.51&0.50&0.69&0.73& \textbf{0.76}&\textbf{0.76}&0.75\\ 
              \hline
              \multirow{3}{2em}{10dB} & PSNR &19.88&23.55&22.47&22.62&18.96&25.43&21.64&31.72&\textbf{32.15}&32.12&31.04&31.24\\ 
              & SSIM &0.50&0.83&0.85&0.75&0.81&0.81&0.88&0.95& \textbf{0.96}& \textbf{0.96}&0.95&\textbf{0.96}\\ 
              & IoU &0.59&0.67&0.70&0.61&0.68&0.66&0.66&0.86& \textbf{0.91}&0.85&0.86&0.87\\ 
              \hline
              \multirow{3}{2em}{15dB} & PSNR &26.37&27.41&26.18&26.01&19.88&29.56&28.00&37.07&37.10&\textbf{37.19}&34.30&36.16\\ 
              & SSIM &0.84&0.91&0.92&0.83&0.88&0.93&0.96& \textbf{0.98}&\textbf{0.98} & \textbf{0.98}&0.97&\textbf{0.98}\\ 
              & IoU &0.85&0.85&0.83&0.83&0.84&0.84&0.89&0.93&\textbf{0.94} &0.93&0.92&\textbf{0.94}\\ 
              \hline
              \multirow{3}{2em}{18dB} & PSNR &31.44&29.52&28.17&30.70&20.23&31.32&31.39&37.97&38.02&\textbf{38.36}&35.00&37.32\\ 
              & SSIM &0.93&0.93&0.94&0.92&0.91&0.96&0.97&0.98&0.98& \textbf{0.99}&0.98&\textbf{0.99}\\ 
			  & IoU  &0.91&0.89&0.85&0.88&0.88& 0.91 &0.93&0.93& \textbf{0.94}& \textbf{0.94}&0.92&\textbf{0.94}\\
            \hline	
            \multirow{3}{2em}{20dB}&PSNR&36.23&30.44&28.99&34.57&20.35& 31.53 &32.95&38.81&38.95&\textbf{39.08}&35.35&37.10\\
              &SSIM &0.97&0.93&0.95&0.96&0.93& 0.97 &0.98& \textbf{0.99}&  \textbf{0.99}&\textbf{0.99}&0.98&\textbf{0.99}\\
              &IoU  &0.94&0.90&0.86&0.90&0.91& 0.90 &0.94&0.94&0.94&0.94&0.93&\textbf{0.95}\\
            \hline	
        \end{tabular}
    }
    \label{table:snr_segmentation_results}																
\end{table*}

  \begin{table*}[t!] 														 							
      \centering 														 								
      \caption{Results of \textbf{object segmentation} using \textbf{MobileNetv2} with respect to varying Doppler Shift } 									
      \scalebox{0.8}{    														 						
          \begin{tabular}{ |l|l|l|l|l|l|l|l|l|l|l|l|l|l|l|} 											
              \hline 														 							
              SNR& &No Filter&Mean&Median&BM3D&DenoiseNet& MIMO-UNet &MPRNET Pre &MPRNET&NR-Netv1&NR-Netv2&NR-Netv3&NR-Netv4\\ 
              \hline 														 							
  			\multirow{3}{3em}{100Hz}  &  PSNR  &28.06&27.37&25.91&27.66&19.92&28.85&27.86& \textbf{32.74} &31.83&31.95&30.82&30.90 \\
  			&	   SSIM  &0.90&0.91&0.92&0.90&0.90&0.95&0.96&0.96& \textbf{0.97}& \textbf{0.97}&0.96&\textbf{0.97} \\
  			&	   IoU  &0.87&0.84&0.87&0.86&0.86&0.85&0.91&0.90&0.91&0.90&0.90&\textbf{0.93} \\
  			\hline 														 								
  			\multirow{3}{3em}{300Hz}  &  PSNR  &19.88&23.55&22.47&22.62&18.90&25.40&21.64&31.72& \textbf{32.15}&32.12&31.04&31.23 \\
  			&	   SSIM  &0.50&0.83&0.85&0.75&0.81&0.82&0.88&0.88& \textbf{0.96}& \textbf{0.96}&0.95&\textbf{0.96} \\
  			&	   IoU  &0.59&0.67&0.70&0.61&0.69&0.66&0.66&0.86& \textbf{0.91}&0.85&0.86&0.87 \\
  			\hline 														 								
  			\multirow{3}{3em}{350Hz}  &  PSNR  &21.50&25.37&24.33&24.78&19.41&26.16&23.39&31.17&31.23&\textbf{31.28}&30.35&30.65 \\
  			&	   SSIM  &0.56&0.85&0.88&0.78&0.83&0.84&0.89&0.89& \textbf{0.96}& \textbf{0.96}&0.95&\textbf{0.96} \\
  			&	  IoU  &0.75&0.74&0.77&0.67&0.73&0.68&0.74&0.83&\textbf{0.84}&0.83&0.82&\textbf{0.84} \\
  			\hline 														 								
  			\multirow{3}{3em}{400Hz}  &  PSNR  &21.89&25.23&24.11&24.31&19.29&25.94&23.77&31.51&32.31& \textbf{33.09}&31.67&31.04 \\
  			&	   SSIM  &0.59&0.86&0.88&0.77&0.82&0.86&0.90&0.90&\textbf{0.96}& \textbf{0.96}&\textbf{0.96}&\textbf{0.96} \\
  			&	   IoU  &0.74&0.76&0.79&0.73&0.74&0.71&0.74&0.85&0.87& \textbf{0.93}&0.87&0.88 \\
  			\hline 														 								
  			\multirow{3}{3em}{500Hz}  &  PSNR  &31.08&29.49&27.91&30.50&20.09&30.14&30.57&35.77&35.85& \textbf{36.08}&33.81&34.27 \\
  			&	   SSIM  &0.91&0.92&0.94&0.91&0.90&0.95&0.96&0.96&\textbf{0.98}& \textbf{0.98}&0.97&\textbf{0.98} \\
  			&	   IoU  &0.88&0.87&0.84&0.87&0.87&0.85&  \textbf{0.92} &0.91& \textbf{0.92}&0.91&0.91&0.91 \\
  			\hline 														 								
  			\multirow{3}{3em}{750Hz}  &  PSNR  &21.52&25.07&24.12&23.31&19.43&26.08&23.71&32.52&32.73& \textbf{33.33}&32.02&31.74 \\
  			&	   SSIM  &0.60&0.86&0.88&0.75&0.83&0.86&0.90&0.90&0.96& \textbf{0.97}&0.96&\textbf{0.97} \\
  			&	   IoU  &0.71&0.78&0.80&0.74&0.74&0.72&0.76&  \textbf{0.91} &0.89&0.89&0.88&0.90 \\
  			\hline 			
  		\end{tabular} 		
      } 					
      \label{table:doppler_segmentation_results} 														 
  \end{table*}

 \begin{table*}[t!] 														 							
      \centering 														 								
      \caption{Results of \textbf{object detection} using \textbf{YOLOv5} with respect to varying channel noise SNR} 										
      \scalebox{0.8}{ 														 							
          \begin{tabular}{ |c|c|c|c|c|c|c|c|c|c|c|c|c|c|c|c|c| } 										
              \hline 														 							
              SNR &	&	 No Filter 	&Mean & Median & BM3D & DenoiseNet & MIMO-UNet& MPRNET Pre& MPRNET &NR-Netv1&NR-Netv2&NR-Netv3&NR-Netv4 \\
              \hline 														 							
              \multirow{3}{3em}{1dB} & mAP@.5&0.40&0.47&0.46&0.50&0.51&0.52&0.50&0.53&\textbf{0.54}&\textbf{0.54}&0.53& \textbf{0.54} \\
             &  mAP@.5:.95&0.23&0.32&0.30&0.35&0.35&0.35&0.35& \textbf{0.41}& \textbf{0.41}& \textbf{0.41}&0.40& \textbf{0.41} \\
              \hline 														 							
              \multirow{3}{3em}{2dB}&  mAP@.5&0.43&0.48&0.48&0.51&0.52&0.53&0.52& \textbf{0.56}&0.55&0.55&0.54&0.55 \\
             &  mAP@.5:.95&0.25&0.34&0.32&0.37&0.39&0.38&0.37&0.43&0.43&0.43&0.42& \textbf{0.44} \\
              \hline 														 							
              \multirow{3}{3em}{3dB}&  mAP@.5&0.48&0.50&0.49&0.52&0.56&\textbf{0.57}&0.54& \textbf{0.57}&0.56& \textbf{0.57}&0.55& \textbf{0.57} \\
             & mAP@.5:.95&0.27&0.37&0.35&0.40&0.42&0.44&0.40& \textbf{0.46}&0.45&0.45&0.44& \textbf{0.46} \\
              \hline 														 							
              \multirow{3}{3em}{4dB}&  mAP@.5&0.47&0.52&0.51&0.54&0.56&0.56&0.55& \textbf{0.58}&0.57& \textbf{0.58}&0.56& \textbf{0.58} \\
             &  mAP@.5:.95&0.31&0.41&0.39&0.42&0.44&0.44&0.43& \textbf{0.48}&0.47& \textbf{0.48}&0.46&0.47 \\
              \hline 														 							
              \multirow{3}{3em}{5dB}&  mAP@.5&0.48&0.53&0.54&0.55&0.57&0.59&0.56&0.59&0.58& \textbf{0.6}&0.58&0.59 \\
             &  mAP@.5:.95&0.32&0.42&0.42&0.44&0.45&0.45&0.47&0.48&0.48& \textbf{0.49}&0.48&0.48 \\
              \hline 														 							
              \multirow{3}{3em}{6dB}&  mAP@.5&0.50&0.54&0.54&0.56&0.58&0.57&0.58& \textbf{0.60}& \textbf{0.60}& \textbf{0.60}&0.58&0.59 \\
             &  mAP@.5:.95&0.35&0.44&0.43&0.45&0.46&0.46&0.47& \textbf{0.51}&0.50&0.50&0.48&0.49 \\
              \hline 														 							
              \multirow{3}{3em}{10dB}&  mAP@.5&0.54&0.59&0.60&0.59&0.58&0.58&0.60& \textbf{0.61}& \textbf{0.61}& \textbf{0.61}&0.58& \textbf{0.61} \\
             &  mAP@.5:.95&0.44&0.51&0.51&0.51&0.47&0.47&0.51& \textbf{0.54}&0.53& \textbf{0.54}&0.51&0.53\\
              \hline 														 							
              \multirow{3}{3em}{15dB}&  mAP@.5&0.59& \textbf{0.61}& \textbf{0.61}&0.59&0.60& \textbf{0.61}& \textbf{0.61}& \textbf{0.61}& \textbf{0.61}& \textbf{0.61}&0.59& \textbf{0.61} \\
             &  mAP@.5:.95&0.51&0.53&0.53&0.51&0.52&0.52& \textbf{0.54}&0.53& \textbf{0.54}& \textbf{0.54}&0.52&0.53 \\
              \hline 														 							
              \multirow{3}{3em}{18dB}&  mAP@.5&0.61& \textbf{0.62}& \textbf{0.62}&0.61&0.61&0.61& \textbf{0.62}&0.61&0.61&0.61&0.59&0.61 \\
             &  mAP@.5:.95&0.53& \textbf{0.54}& \textbf{0.54}&0.53&0.53&0.53& \textbf{0.54}&0.53&0.53&0.53&0.52& \textbf{0.54} \\
              \hline 														 							
              \multirow{3}{3em}{20dB}&  mAP@.5& \textbf{0.62}&0.61&0.61&0.61& \textbf{0.62}& \textbf{0.62}&0.61&0.61&0.61&0.61&0.59&0.61 \\
             &  mAP@.5:.95& \textbf{0.54}& \textbf{0.54}& \textbf{0.54}& \textbf{0.54}& \textbf{0.54}& \textbf{0.54}& \textbf{0.54}& \textbf{0.54}& \textbf{0.54}& \textbf{0.54}&0.52& \textbf{0.54} \\
              \hline 														 							
          \end{tabular} 														 						
      } 														 										
      \label{table:snr_detection_results} 														 		
  \end{table*} 														 													
 \begin{table*}[t!]
  	\centering
      \caption{Results of \textbf{object detection} using \textbf{YOLOv5} with respect to varying Doppler shift}
      \scalebox{0.8}{
          \begin{tabular}{ |c|c|c|c|c|c|c|c|c|c|c|c|c|c|}
              \hline
              SNR & & No Filter& Mean& Median& BM3D&DenoiseNet&MIMO-UNet&MPRNET Pre&MPRNET&NR-Netv1&NR-Netv2&NR-Netv3&NR-Netv4 \\
              \hline
             \multirow{3}{3em}{100Hz}& mAP@.5&0.61&0.61&0.61&0.61&\textbf{0.62}&\textbf{0.62}&\textbf{0.62}&0.61&0.61&0.61&0.59&0.60 \\
             & mAP@.5:.95&\textbf{0.54}&0.53&\textbf{0.54}&\textbf{0.54}&\textbf{0.54}&\textbf{0.54}&\textbf{0.54}&0.53&0.53&0.53&0.51&0.53 \\
              \hline
              \multirow{3}{3em}{300Hz} & mAP@.5 &0.54&0.60&0.60&0.59&0.58&0.59&0.60&\textbf{0.61}&\textbf{0.61}&\textbf{0.61}&0.58&\textbf{0.61} \\
              & mAP@.5:.95&0.44&0.51&0.51&0.51&0.47&0.50&0.51&\textbf{0.54}&0.53&\textbf{0.54}&0.51&0.53 \\
              \hline
              \multirow{3}{3em}{350Hz} & mAP@.5 &0.55&0.59&0.60&0.56&0.59&0.60&\textbf{0.61}&\textbf{0.61}&\textbf{0.61}&\textbf{0.61}&0.60&\textbf{0.61} \\
              & mAP@.5:.95&0.45&0.51&\textbf{0.53}&0.47&0.49&0.52&\textbf{0.53}&\textbf{0.53}&\textbf{0.53}&\textbf{0.53}&0.51&\textbf{0.53} \\
              \hline
              \multirow{3}{3em}{400Hz} & mAP@.5 &0.57&0.61&\textbf{0.62}&0.60&0.59&0.58&0.61&0.61&0.60&0.61&0.60&0.61 \\
              & mAP@.5:.95&0.48&0.53&0.53&0.51&0.50&0.50&0.52&0.53&0.53&\textbf{0.54}&0.52&\textbf{0.54} \\
              \hline
              \multirow{3}{3em}{500Hz} & mAP@.5 &0.60&\textbf{0.62}&\textbf{0.62}&0.60&0.61&0.61&0.61&0.61&0.61&0.61&0.59&0.61 \\
              & mAP@.5:.95&0.52&\textbf{0.54}&\textbf{0.54}&0.52&0.53&0.53&\textbf{0.54}&0.53&\textbf{0.54}&\textbf{0.54}&0.52&0.53 \\
              \hline
              \multirow{3}{3em}{750Hz} & mAP@.5 &0.56&0.61&0.61&0.58&0.59&0.61&0.61&\textbf{0.62}&\textbf{0.62}&0.61&0.60&\textbf{0.62} \\
              & mAP@.5:.95&0.46&0.52&0.53&0.48&0.49&0.52&0.53&\textbf{0.54}&\textbf{0.54}&0.53&0.52&0.53 \\
              \hline          
  		\end{tabular}
      }
  	\label{table:doppler_detection_results}
\end{table*}

Here, we study (i) how the SNR, which characterizes the effect of the distance and the obstacles along the path between the UAV and the gNB, and (ii) how the Doppler shift, which represents the speed of the UAV since we assume our gNB is stationary, affects the performance of a task specific (off-the-shelf) and pretrained deep network on the edge server. Table \ref{table:parameters} summarizes our results obtained at various SNR and Doppler shift values.

Fig. \ref{fig:detection_res} demonstrates our results on how changing the SNR and Doppler shift values affect the segmentation and detection results of MobileNetv2 and YOLOv5. In the figure, the red line represents the best value (the value obtained when there is no noise introduced by the communication system). The blue line represents the output of the off-the-shelf algorithms at different SNR values (in Fig. \ref{fig:iou_snr} and Fig. \ref{fig:map_snr}) and at different Doppler shift values (in Fig. \ref{fig:iou_doppler} and Fig. \ref{fig:map_doppler}). As demonstrated in Fig. \ref{fig:iou_snr}, the segmentation algorithm yields better results, as the SNR value increases. However, Doppler effect is relatively smaller, which indicates that 5G system is better at eliminating the effect of the Doppler shift. On the other hand, our preliminary results performed on a 4G system (see \cite{ilhan2021offloading}) indicated that a 4G system's performance is more susceptible to Doppler shift. Fig. \ref{fig:map_snr} and Fig. \ref{fig:map_doppler} summarize our results on how changing the SNR and Doppler shift values affect the object detection results of the YOLOv5 algorithm. Similar to segmentation, for object detection, the detection performance increases as the SNR value increases by converging to the case where there is no noise. Next, we study the performance of various denoising algorithms to eliminate the effect of the noise.

\subsubsection{The effect of using a denoiser on the edge server}

As demonstrated in the previous subsection, some SNR values can affect the performance of a pretrained network significantly (even the Doppler effect can be considered an important effect in certain deep applications). The performance can drop drastically (where mIoU can drop yielding a 82\% less performance, see the mIoU drop from 0.94 to 0.12 in Fig. \ref{fig:iou_snr}) making the pretrained algorithm less reliable. Consequently, a denoising stage is necessary. Considering the data is a spatial data (i.e., image data), spatial denoising algorithms can be used here. Therefore, we propose using a denoising stage at the edge server, prior to using a pretrained network when a deep learning based task is offloaded. In our denosing stage, we compare the performance of various algorithms, from classical algorithms to recently proposed deep approaches. In this part, we use various denoising algorithms including mean filtering, median filtering, BM3D, DenoiseNet, MPRNET, MIMO-Unet, in addition to our NR-Net. Table \ref{table:snr_segmentation_results} and  Table \ref{table:doppler_segmentation_results} summarizes our experimental results obtained for pretrained MobileNetv2 for various SNR values and for various Doppler shift values, respectively. In the tables, we use three metrics: PSNR, SSIM and IoU. The best values are shown in bold. In the tables, "MPRNET Pre" column shows the results of pretrained MPRNET and the "MPRNET" column shows the results of MPRNET after being re-trained on the dataset. Table \ref{table:snr_detection_results} (for various SNR values) and Table \ref{table:doppler_detection_results} (for various Doppler shift values) summarize our experimental results obtained on pretrained YOLOv5 in mAP value. In the tables, "no filter" represents the case where there is no algorithm is used for the denoising block. Fig. \ref{fig:result} shows qualitative results.

Since the variance between neighbour pixels can be higher at low SNR values, an average value is produced after applying the mean filter and noisy pixels are smoothed out. However, when there is negligible noise in the data passed through the wireless channel with high SNR values, it is observed that the images with a mean filter produced worse results than the images without any filter. Similarly, in the case of the median filter, the images passed through the low SNR channel yield better results than the unfiltered images, while images passed through the high SNR channel yields lower performance when compared to the case without using any filter. We used 5x5 filter in both mean and median filters.
Since BM3D was put forth with an adaptive filtering strategy, it achieves more successful results than more basic approaches such as mean and median filters. When the noise level in the channel is low, better IoU and PSNR results are obtained with BM3D than the results of both mean and median filters. Since it is based on the BM3D Block Matching principle, the IoU results cannot be seen to increase with PSNR due to the smoothing of the segmented areas in low-noise images.
DenoiseNet was trained from scratch with our dataset over 80 epochs.
In our experiments, all the used deep architectures are trained from scratch on the same training data. Then we checked their performance on the test dataset where we used the output of each denoising algorithm to both off-the-shelf pretrained deep algorithms: MobileNetV2 and YOLOv5.

\section{Discussion and Conclusion}

In this paper, we first described and analyzed a problem caused due to the noise which is introduced by the used wireless communication system, when a deep-learning-based task is offloaded. Then, we introduced a novel deep learning based solution: Noise-Removing Net (NR-Net) to handle such noise, when deep learning based object segmentation and object detection tasks are offloaded to a remote (edge) server. In particular: (i) we identified an important issue on offloading a deep learning based task to a remote edge server, when a noisy fading wireless 5G system is used; (ii) we designed a simulation environment to investigate how the performance of the edge computing system varies with the distributive effects of the communication channel. As our variables, we picked Doppler shift caused by the UAV velocity and the noise power caused by the environment properties such as the distance between transmitter and receiver, obstacles and weather conditions. Finally, (iii) we introduced a novel and state-of-the-art denoising algorithm that helps gaining better results, when pretrained off-the-shelf deep algorithms are used on the edge server.

Our analysis shows that offloading over a wireless channel affects the performance of  both used pretrained deep-learning-based semantic image segmentation and object detection algorithms. That indicates that the pretrained segmentation and detection models are sensitive to channel noise. Therefore, we suggest that a denoiser algorithm should be used at the receiver side to increase the performance of such networks and to make them more robust with respect to the channel noise. We compared 6 different denoising algorithms (from basics to deep-learning based advances) to our proposed NR-Net. We report that our proposed deep denoising algorithm: NR-Net yielded the best results in many situations in our experiments (see tables \ref{table:snr_segmentation_results}, \ref{table:doppler_segmentation_results}, \ref{table:snr_detection_results} and \ref{table:doppler_detection_results}). Our proposed NR-Net yielded the best results (on average) especially at lower SNR values. A potential future direction is studying the performance of our framework in 6G systems. Another potential future direction is studying the effect of using additional noise eliminating steps as traditionally used in 5G systems at different layers.

\section*{Acknowledgment}
{\small This paper has been produced benefiting from the 2232 International Fellowship for Outstanding Researchers Program of TÜBİTAK (Project No:118C356). However, the entire responsibility of the paper belongs to the owner of the paper. The financial support received from TÜBİTAK does not mean that the content of the publication is approved in a scientific sense by TÜBİTAK.}

{
\bibliographystyle{IEEEtran}
\bibliography{refs}

\begin{thebibliography}{10}
\providecommand{\url}[1]{#1}
\csname url@samestyle\endcsname
\providecommand{\newblock}{\relax}
\providecommand{\bibinfo}[2]{#2}
\providecommand{\BIBentrySTDinterwordspacing}{\spaceskip=0pt\relax}
\providecommand{\BIBentryALTinterwordstretchfactor}{4}
\providecommand{\BIBentryALTinterwordspacing}{\spaceskip=\fontdimen2\font plus
\BIBentryALTinterwordstretchfactor\fontdimen3\font minus
  \fontdimen4\font\relax}
\providecommand{\BIBforeignlanguage}[2]{{%
\expandafter\ifx\csname l@#1\endcsname\relax
\typeout{** WARNING: IEEEtran.bst: No hyphenation pattern has been}%
\typeout{** loaded for the language `#1'. Using the pattern for}%
\typeout{** the default language instead.}%
\else
\language=\csname l@#1\endcsname
\fi
#2}}
\providecommand{\BIBdecl}{\relax}
\BIBdecl

\bibitem{albaba2020synet}
B.~M. Albaba and S.~Ozer, ``{SyNet}: An ensemble network for object detection
  in {UAV} images,'' in \emph{International Conference on Pattern Recognition
  (ICPR2020)}, 2020.

\bibitem{sahin2021yolodrone}
O.~Sahin and S.~Ozer, ``Yolodrone: Improved yolo architecture for object
  detection in drone images,'' in \emph{2021 44th International Conference on
  Telecommunications and Signal Processing (TSP)}.\hskip 1em plus 0.5em minus
  0.4em\relax IEEE, 2021, pp. 361--365.

\bibitem{gozenvisual}
D.~G{\"o}zen and S.~Ozer, ``{Visual Object Tracking in Drone Images with Deep
  Reinforcement Learning},'' in \emph{International Conference on Pattern
  Recognition (ICPR2020)}, 2020.

\bibitem{ozer2022siamesefuse}
S.~Ozer, M.~Ege, and M.~A. {\"O}zkanoglu, ``Siamesefuse: A computationally
  efficient and a not-so-deep network to fuse visible and infrared images,''
  \emph{Pattern Recognition}, vol. 129, p. 108712, 2022.

\bibitem{ozkanoglu2022infragan}
M.~A. {\"O}zkano{\u{g}}lu and S.~Ozer, ``Infragan: A gan architecture to
  transfer visible images to infrared domain,'' \emph{Pattern Recognition
  Letters}, vol. 155, pp. 69--76, 2022.

\bibitem{huang2018yolo}
R.~Huang, J.~Pedoeem, and C.~Chen, ``Yolo-lite: a real-time object detection
  algorithm optimized for non-gpu computers,'' in \emph{2018 IEEE International
  Conference on Big Data (Big Data)}.\hskip 1em plus 0.5em minus 0.4em\relax
  IEEE, 2018, pp. 2503--2510.

\bibitem{sandler2018mobilenetv2}
M.~Sandler, A.~Howard, M.~Zhu, A.~Zhmoginov, and L.-C. Chen, ``{MobileNetV2}:
  Inverted residuals and linear bottlenecks,'' in \emph{Proceedings of the IEEE
  Conference on Computer Vision and Pattern Recognition}, 2018, pp. 4510--4520.

\bibitem{chen2020intelligent}
J.~Chen, S.~Chen, S.~Luo, Q.~Wang, B.~Cao, and X.~Li, ``An intelligent task
  offloading algorithm (itoa) for uav edge computing network,'' \emph{Digital
  Communications and Networks}, vol.~6, no.~4, pp. 433--443, 2020.

\bibitem{mukherjee2020distributed}
M.~Mukherjee, V.~Kumar, A.~Lat, M.~Guo, R.~Matam, and Y.~Lv, ``Distributed deep
  learning-based task offloading for uav-enabled mobile edge computing,'' in
  \emph{IEEE INFOCOM 2020-IEEE Conference on Computer Communications Workshops
  (INFOCOM WKSHPS)}.\hskip 1em plus 0.5em minus 0.4em\relax IEEE, 2020, pp.
  1208--1212.

\bibitem{zha2019rank}
Z.~Zha, X.~Yuan, B.~Wen, J.~Zhou, J.~Zhang, and C.~Zhu, ``From rank estimation
  to rank approximation: Rank residual constraint for image restoration,''
  \emph{IEEE Transactions on Image Processing}, vol.~29, pp. 3254--3269, 2019.

\bibitem{tian2020deep}
C.~Tian, L.~Fei, W.~Zheng, Y.~Xu, W.~Zuo, and C.-W. Lin, ``Deep learning on
  image denoising: An overview,'' \emph{Neural Networks}, vol. 131, pp.
  251--275, 2020.

\bibitem{glenn_jocher_2020_4154370}
\BIBentryALTinterwordspacing
G.~Jocher, A.~Stoken, J.~Borovec, NanoCode012, ChristopherSTAN, L.~Changyu,
  Laughing, tkianai, A.~Hogan, lorenzomammana, yxNONG, AlexWang1900,
  L.~Diaconu, Marc, wanghaoyang0106, ml5ah, Doug, F.~Ingham, Frederik, Guilhen,
  Hatovix, J.~Poznanski, J.~Fang, L.~Yu, changyu98, M.~Wang, N.~Gupta,
  O.~Akhtar, PetrDvoracek, and P.~Rai, ``{ultralytics/yolov5: v6.0 - YOLOv5n
  'Nano' models, Roboflow integration, TensorFlow export, OpenCV DNN
  support},'' Oct. 2021. [Online]. Available:
  \url{https://doi.org/10.5281/zenodo.5563715}
\BIBentrySTDinterwordspacing

\bibitem{ilhan2021offloading}
H.~E. Ilhan, S.~Ozer, G.~K. Kurt, and H.~A. Cirpan, ``Offloading deep learning
  empowered image segmentation from uav to edge server,'' in \emph{2021 44th
  International Conference on Telecommunications and Signal Processing
  (TSP)}.\hskip 1em plus 0.5em minus 0.4em\relax IEEE, 2021, pp. 296--300.

\bibitem{kunst2020application}
R.~Kunst, E.~Pignaton, T.~Zhou, and H.~Hu, ``Application of future 6g
  technology to support heavy data traffic in highly mobile networks,'' in
  \emph{2020 First International Conference of Smart Systems and Emerging
  Technologies (SMARTTECH)}.\hskip 1em plus 0.5em minus 0.4em\relax IEEE, 2020,
  pp. 144--148.

\bibitem{9622148}
B.~Dai, J.~Niu, T.~Ren, Z.~Hu, and M.~Atiquzzaman, ``Towards energy-efficient
  scheduling of uav and base station hybrid enabled mobile edge computing,''
  \emph{IEEE Transactions on Vehicular Technology}, vol.~71, no.~1, pp.
  915--930, 2022.

\bibitem{zeng2017energy}
Y.~Zeng and R.~Zhang, ``Energy-efficient {UAV} communication with trajectory
  optimization,'' \emph{IEEE Transactions on Wireless Communications}, vol.~16,
  no.~6, pp. 3747--3760, 2017.

\bibitem{callegaro2018optimal}
D.~Callegaro and M.~Levorato, ``Optimal computation offloading in edge-assisted
  {UAV} systems,'' in \emph{2018 IEEE Global Communications Conference
  (GLOBECOM)}.\hskip 1em plus 0.5em minus 0.4em\relax IEEE, 2018, pp. 1--6.

\bibitem{li2018task}
J.~Li, Q.~Liu, P.~Wu, F.~Shu, and S.~Jin, ``Task offloading for {UAV}-based
  mobile edge computing via deep reinforcement learning,'' in \emph{IEEE/CIC
  International Conference on Communications in China (ICCC)}, 2018, pp.
  798--802.

\bibitem{8210823}
C.~{Ting}, X.~{Yun}, Z.~{Xiangmo}, G.~{Tao}, and X.~{Zhigang}, ``{4G} {UAV}
  communication system and hovering height optimization for public safety,'' in
  \emph{IEEE International Conference on e-Health Networking, Applications and
  Services (Healthcom)}, 2017, pp. 1--6.

\bibitem{costanzo2020dynamic}
F.~Costanzo, P.~Di~Lorenzo, and S.~Barbarossa, ``Dynamic resource optimization
  and altitude selection in {UAV-}based multi-access edge computing,'' in
  \emph{IEEE International Conference on Acoustics, Speech and Signal
  Processing (ICASSP)}, 2020, pp. 4985--4989.

\bibitem{kim2019optimal}
K.~Kim and C.~S. Hong, ``Optimal task-{UAV}-edge matching for computation
  offloading in {UAV} assisted mobile edge computing,'' in \emph{Asia-Pacific
  Network Operations and Management Symposium}, 2019, pp. 1--4.

\bibitem{kim2020machine}
K.~Kim, Y.~M. Park, and C.~S. Hong, ``Machine learning based edge-assisted
  {UAV} computation offloading for data analyzing,'' in \emph{International
  Conference on Information Networking (ICOIN)}, 2020.

\bibitem{liu2020incentive}
Y.~Liu, M.~Qiu, J.~Hu, and H.~Yu, ``Incentive {UAV} enabled mobile edge
  computing based on microwave power transmission,'' \emph{IEEE Access},
  vol.~8, pp. 28\,584--28\,593, 2020.

\bibitem{zhang2019edge}
Q.~Zhang, H.~Sun, X.~Wu, and H.~Zhong, ``Edge video analytics for public
  safety: A review,'' \emph{Proceedings of the IEEE}, vol. 107, no.~8, pp.
  1675--1696, 2019.

\bibitem{remez2017deep}
T.~Remez, O.~Litany, R.~Giryes, and A.~M. Bronstein, ``Deep convolutional
  denoising of low-light images,'' \emph{arXiv preprint arXiv:1701.01687},
  2017.

\bibitem{everingham2015pascal}
M.~Everingham, S.~A. Eslami, L.~Van~Gool, C.~K. Williams, J.~Winn, and
  A.~Zisserman, ``The pascal visual object classes challenge: A
  retrospective,'' \emph{International Journal of Computer Vision}, vol. 111,
  no.~1, pp. 98--136, 2015.

\bibitem{DBLP:journals/corr/abs-2102-02808}
\BIBentryALTinterwordspacing
S.~W. Zamir, A.~Arora, S.~H. Khan, M.~Hayat, F.~S. Khan, M.~Yang, and L.~Shao,
  ``Multi-stage progressive image restoration,'' \emph{CoRR}, vol.
  abs/2102.02808, 2021. [Online]. Available:
  \url{https://arxiv.org/abs/2102.02808}
\BIBentrySTDinterwordspacing

\bibitem{cho2021rethinking}
S.-J. Cho, S.-W. Ji, J.-P. Hong, S.-W. Jung, and S.-J. Ko, ``Rethinking
  coarse-to-fine approach in single image deblurring,'' 2021.

\bibitem{patel2018comparative}
S.~Patel, V.~Shah, and M.~Kansara, ``Comparative study of 2g, 3g and 4g,''
  \emph{International Journal of Scientific Research in Computer Science,
  Engineering and Information Technology}, vol.~3, no.~3, pp. 1962--1964, 2018.

\bibitem{8412469}
A.~A. Zaidi, R.~Baldemair, V.~Moles-Cases, N.~He, K.~Werner, and A.~Cedergren,
  ``Ofdm numerology design for 5g new radio to support iot, embb, and mbsfn,''
  \emph{IEEE Communications Standards Magazine}, vol.~2, no.~2, pp. 78--83,
  2018.

\bibitem{3gpp5g}
\emph{Multiplexing and channel coding}, 3GPP, 3 2018, version 15.2.0 Release
  15.

\bibitem{gallager1962low}
R.~Gallager, ``Low-density parity-check codes,'' \emph{IRE Transactions on
  information theory}, vol.~8, no.~1, pp. 21--28, 1962.

\bibitem{3gpp38901}
\emph{Study on channel model for frequencies from 0.5 to 100 GHz}, 3GPP, 7
  2017, version 14.0.0 Release 14.

\bibitem{fan2019brief}
L.~Fan, F.~Zhang, H.~Fan, and C.~Zhang, ``Brief review of image denoising
  techniques,'' \emph{Visual Computing for Industry, Biomedicine, and Art},
  vol.~2, no.~1, pp. 1--12, 2019.

\bibitem{dabov2007image}
K.~Dabov, A.~Foi, V.~Katkovnik, and K.~Egiazarian, ``Image denoising by sparse
  3-d transform-domain collaborative filtering,'' \emph{IEEE Transactions on
  image processing}, vol.~16, no.~8, pp. 2080--2095, 2007.

\bibitem{liu2021Swin}
Z.~Liu, Y.~Lin, Y.~Cao, H.~Hu, Y.~Wei, Z.~Zhang, S.~Lin, and B.~Guo, ``Swin
  transformer: Hierarchical vision transformer using shifted windows,''
  \emph{International Conference on Computer Vision (ICCV)}, 2021.

\bibitem{layernorm}
J.~L. Ba, J.~R. Kiros, and G.~E. Hinton, ``Layer normalization,'' \emph{arXiv
  preprint arXiv:1607.06450}, 2016.

\bibitem{attentionisallyouneed}
A.~Vaswani, N.~Shazeer, N.~Parmar, J.~Uszkoreit, L.~Jones, A.~N. Gomez, L.~u.
  Kaiser, and I.~Polosukhin, ``Attention is all you need,'' in \emph{Advances
  in Neural Information Processing Systems}, I.~Guyon, U.~V. Luxburg,
  S.~Bengio, H.~Wallach, R.~Fergus, S.~Vishwanathan, and R.~Garnett, Eds.,
  vol.~30.\hskip 1em plus 0.5em minus 0.4em\relax Curran Associates, Inc.,
  2017.

\bibitem{li2020fss}
X.~Li, T.~Wei, Y.~P. Chen, Y.-W. Tai, and C.-K. Tang, ``Fss-1000: A 1000-class
  dataset for few-shot segmentation,'' in \emph{Conference on Computer Vision
  and Pattern Recognition}, 2020.

\bibitem{lin2015microsoft}
T.-Y. Lin, M.~Maire, S.~Belongie, L.~Bourdev, R.~Girshick, J.~Hays, P.~Perona,
  D.~Ramanan, C.~L. Zitnick, and P.~Dollár, ``Microsoft coco: Common objects
  in context,'' 2015.

\bibitem{5596999}
A.~Horé and D.~Ziou, ``Image quality metrics: Psnr vs. ssim,'' in \emph{2010
  20th International Conference on Pattern Recognition}, 2010, pp. 2366--2369.

\bibitem{1284395}
Z.~Wang, A.~Bovik, H.~Sheikh, and E.~Simoncelli, ``Image quality assessment:
  from error visibility to structural similarity,'' \emph{IEEE Transactions on
  Image Processing}, vol.~13, no.~4, pp. 600--612, 2004.

\bibitem{10.1007/978-3-642-40994-3_29}
K.~Boyd, K.~H. Eng, and C.~D. Page, ``Area under the precision-recall curve:
  Point estimates and confidence intervals,'' in \emph{Machine Learning and
  Knowledge Discovery in Databases}, H.~Blockeel, K.~Kersting, S.~Nijssen, and
  F.~{\v{Z}}elezn{\'y}, Eds.\hskip 1em plus 0.5em minus 0.4em\relax Berlin,
  Heidelberg: Springer Berlin Heidelberg, 2013, pp. 451--466.

\end{thebibliography}
}

\par
\section*{Authors' Bio:}
 \textbf{Sedat Ozer} received his M.Sc. degree from Univ. of Massachusetts, Dartmouth and his Ph.D. degree from Rutgers University, NJ. He has worked as a research associate in various institutions including Univ. of Virginia and Massachusetts Institute of Technology. His research interests include pattern analysis, object detection \& segmentation, object tracking, visual data analysis, geometric and explainable AI algorithms and explainable fusion algorithms. As a recipient of TUBITAK's international outstanding research fellow and as an Assistant Professor, he is currently at the department of Computer Science at Ozyegin University.

\par
\par
  \textbf{Huseyin Enes Ilhan} received his B.Eng. degree from Marmara University, Turkey, in 2019. He is currently a M.Sc. student at Istanbul Technical University. His current research interests are telecommunication systems, image processing and artificial intelligence.

\par
\par
  \textbf{Mehmet Akif Özkanoğlu} is currently a M.Sc. student at Bilkent University and is working on designing efficient algorithms for autonomous systems. He received his B.Sc. degree from Istanbul Technical University and his research interests include deep learning, object detection, tracking and robotics. 
 \par 
\par
  \textbf{Hakan Ali Cirpan} (Member, IEEE) received the B.S. degree from Uludag University, Bursa, Turkey, in 1989, the M.S. degree from the University of Istanbul, Istanbul, Turkey, in 1992, and the Ph.D. degree from the Stevens Institute of Technology, Hoboken, NJ, USA, in 1997, all in electrical engineering. From 1995 to 1997, he was a Research Assistant with the Stevens Institute of Technology, Hoboken, NJ, USA, working on signal processing algorithms for wireless communication systems. In 1997, he joined the faculty of the Department of Electrical and Electronics Engineering at The University of Istanbul. In 2010, he has joined to the faculty of the Department of Electronics and Communication Engineering at Istanbul Technical University. His general research interests cover wireless communications, statistical signal and array processing, system identification, and estimation theory. His current research activities are focused on machine learning, signal processing, and communication concepts with specific attention to next generation mobile wireless communication systems. Dr. Cirpan is a member of Sigma Xi. He was the recipient of the Peskin Award from Stevens Institute of Technology as well as Prof. Nazim Terzioglu Award from the Research Fund of The University of Istanbul.

\newpage

\end{document}